\documentclass[runningheads]{llncs}

\usepackage{amsmath,amsfonts,bm}
\usepackage{xfrac}
\usepackage{pifont}

\def\eqref#1{equation~\ref{#1}}

\def\1{\bm{1}}

\DeclareMathAlphabet{\mathsfit}{\encodingdefault}{\sfdefault}{m}{sl}
\SetMathAlphabet{\mathsfit}{bold}{\encodingdefault}{\sfdefault}{bx}{n}

\newcommand{\xmark}{\ding{55}}%

\def\bs{\boldsymbol}

\def\mc{\mathcal}
\def\tt{\texttt}

\def\eg{\textit{e.g.}}
\def\ie{\textit{i.e.}}

\newcommand{\mathshift}[4][3pt]{
        \text{\raisebox{#1}{\smash{\fontsize{#2}{#3}$#4$}}}
        }
 
\usepackage{eccv}

\usepackage{eccvabbrv}

\usepackage{graphicx}
\usepackage{booktabs}

\usepackage[accsupp]{axessibility}  

\usepackage{microtype}

\usepackage{amsmath}
\usepackage{amssymb}
\usepackage{mathtools}
\usepackage{url}            
\usepackage{booktabs}       
\usepackage{amsfonts}       
\usepackage{nicefrac}       
\usepackage{microtype}      
\usepackage{xcolor}         
\usepackage{wrapfig}
\usepackage{graphicx}
\usepackage{bbding}
\usepackage{algorithm}
\usepackage{algorithmic}
\usepackage{multirow}
\usepackage{diagbox}
\usepackage{bm}
\usepackage{colortbl}
\usepackage{scalerel}
\usepackage{color}
\usepackage{listings}
\definecolor{dkgreen}{rgb}{0,0.6,0}
\definecolor{gray}{rgb}{0.5,0.5,0.5}
\definecolor{mauve}{rgb}{0.58,0,0.82}
\definecolor{dgreen}{RGB}{0, 102, 0}
\definecolor{flesh}{RGB}{255,229,217}
\definecolor{lgray}{RGB}{236, 236, 236}

\usepackage{hyperref}

\usepackage{orcidlink}

\begin{document}

\title{Unsqueeze \texttt{[CLS]} Bottleneck to Learn Rich Representations} 

\titlerunning{Unsqueeze \texttt{[CLS]} Bottleneck to Learn Rich Representations}

\author{Qing Su, and Shihao Ji\thanks{Authors' email addresses: \texttt{\{qsu3, sji\}@gsu.edu}}}

\authorrunning{Q. Su, S. Ji}

\institute{Department of Computer Science\\Georgia State University, USA}

\maketitle

\begin{abstract}
Distillation-based self-supervised learning typically leads to more compressed representations due to its radical clustering process and the implementation of a sharper target distribution. To overcome this limitation and preserve more information from input, we introduce \textbf{UDI}, conceptualized as \textbf{U}nsqueezed \textbf{Di}stillation-based self-supervised learning (SSL). UDI enriches the learned representation by encouraging multimodal prediction distilled from a consolidated profile of local predictions that are derived via stratified sampling. Our evaluations show that UDI not only promotes semantically meaningful representations at instance level, delivering superior or competitive results to state-of-the-art SSL methods in image classification, but also effectively preserves the nuisance of input, which yields significant improvement in dense prediction tasks, including object detection and segmentation. Additionally, UDI performs competitively in low-shot image classification, improving the scalability of joint-embedding pipelines. Various visualizations and ablation studies are presented to further elucidate the mechanisms behind UDI. Our source code is available at \url{https://github.com/ISL-CV/udi}. 

  \vspace{-2mm}
  \vspace{-3mm}
\end{abstract}

\section{Introduction}
\label{sec:intro}
\vspace{-2mm}
Self-supervised learning (SSL) derives representations exclusively from images~\cite{caron2018deep, simclr, mae, dino, ibot}, mirroring the pretext tasks (e.g., language modeling) in NLP domain. This approach aims to train models to understand the intrinsic properties of visual data without explicit external labeling. As highlighted in DINOv2~\cite{oquab2023dinov2}, SSL is capable of capturing a spectrum of information within an image, across various levels of abstraction. Seminal works such as MoCov3~\cite{mocov3}, DINO~\cite{dino}, and iBOT~\cite{ibot} demonstrate that SSL not only yields linear evaluation performance comparable to supervised learning but also significantly enhances performance in fine-tuning and transfer learning, promoting semantically meaningful representations. 

\begin{figure}[tb]
  \centering
  \begin{minipage}[c]{0.74\textwidth}
  \vspace{2mm}
  \includegraphics[width=\textwidth]{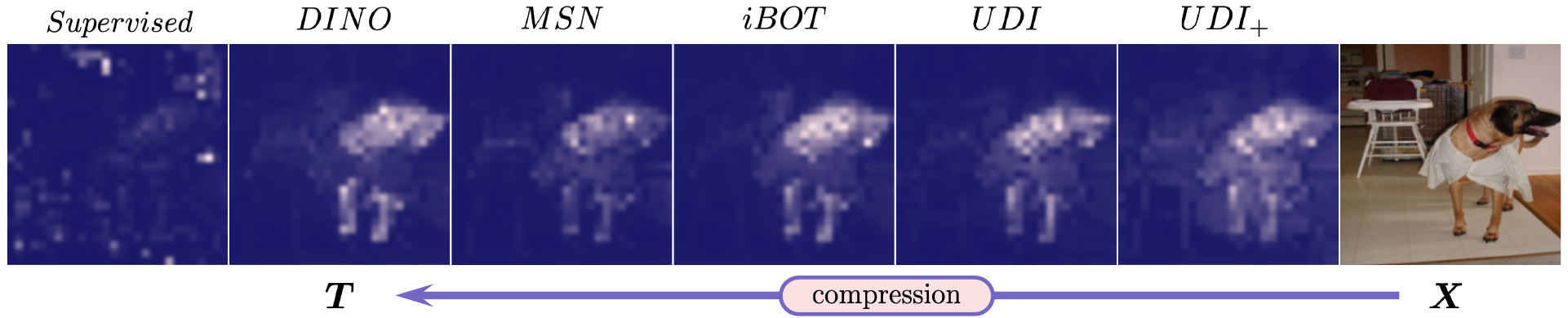}\vspace{-1mm}
  \caption{\footnotesize{\bf Information compression}. UDI retains the most information from $\bs{X}$ as its attention map of $\texttt{CLS}+$ token (UDI$_+$) closely aligned with the underlying semantics.}
  \label{fig:selfattn}
  \end{minipage}
  \begin{minipage}[c]{0.24\textwidth}
  \includegraphics[width=\textwidth]{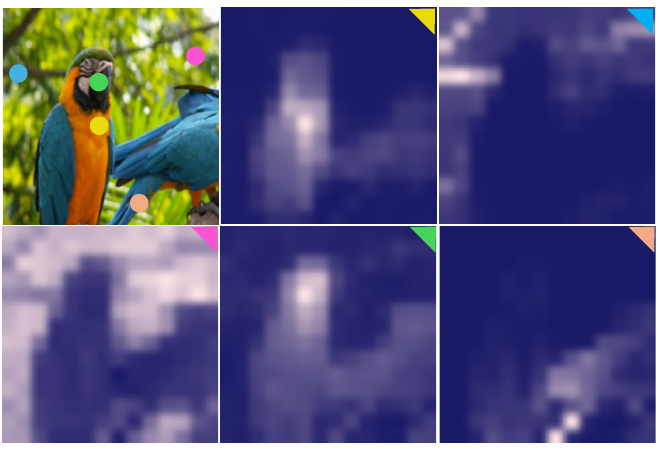}\vspace{-2mm}
  \caption{\footnotesize{\bf Semantic constraints} of SA layer in SRS module.}
  \label{fig:semantic_constraint}
  \end{minipage}
  \vspace{-7mm}
\end{figure}


\indent Yet, existing SSL methods still fall short of the "goal" to capture a comprehensive spectrum of information within an image\cite{oquab2023dinov2}. Specifically, \textbf{implicit clustering} driven by the training objectives, such as non-contrastive loss (\eg, VICReg~\cite{vicreg}) and contrastive loss (\eg, SimCLR~\cite{simclr}), leads to weak compression, potentially resulting in data overfitting~\cite{tschannen2019mutual}. Conversely, \textbf{explicit clustering} that underpins methods, such as self-distillation-based (\eg, DINO, iBOT) and K-means-based methods (\eg, SwAV~\cite{swav}, MSN~\cite{msn}), emphasize more control over clustering to achieve semantically meaningful image abstractions. However, as illustrated in Fig.~\ref{fig:selfattn}, these methods encounter issues of over-compression mainly as a result of employing skewed sharpening techniques, which, while enhancing semantic clarity, can also lead to loss of meaningful nuisances. Moreover, most of these methods consider only the representation of a whole image, limiting the learned semantics to image level. Recent endeavors have intuitively integrated image-level loss with objectives at finer granularities, such as small crops~\cite{detco, soco, orl}, blocks~\cite{resim, scrl}, patches~\cite{densecl, selfpatch, esvit}, and regions~\cite{croc, detcon}. Nonetheless, these methods primarily focus on improving performance in dense prediction tasks and pay less attention to their impact on the quality of image-level representations. In addition, there is a lack of in-depth analysis on how objectives at different levels interact with each other and their impact on performance in downstream tasks.

We further investigate the issue of semantic misalignment in current approaches that utilize multi-level training objectives. The misalignment can be attributed to two main factors: (i) the discrepancy between the imposed semantic constraints and the actual semantic layout of a natural image; for instance, the patch-wise contrastive loss employed by DenseCL\cite{densecl} and DetCo\cite{detco} encourages patch representations to be distinct from the rest of the image, which often contradicts the nature of image semantics and yields limited improvements in downstream tasks; and (ii) enforced alignment of the semantics from different levels, as exemplified by iBOT's implementation of a shared projector for both patch prediction and global representations, as noted in DINOv2~\cite{oquab2023dinov2}. The enforced uniformity disregards the difference between the patch-level contextual information learned via masked image modeling~\cite{mae, sit} and image-level semantics.

In light of the above observations, we propose a novel method by initially identifying the constraints of current SSL approaches through the lens of \emph{Multiview Information Bottleneck} (MIB). By demonstrating the critical role of different compression processes in shaping the quality of the learned representations, we adopt self-distillation as the learning principle across different levels. This is because explicit clustering allows for enhanced control over the clustering process (via the number of centroids, temperatures), which in turn promotes a structured feature space, improving linear performance as demonstrated in many studies~\cite{dino, ibot, oquab2023dinov2, swav, mugs}. To further address the semantic misalignment issue, we extract the patch-level representations through an extra self-attention (SA) update. The SA mechanism~\cite{attention}, functioning as a soft-masked pooling with attention map, offers a semantically coherent constraint that reflects local context, as opposed to the conventional discriminative constraints per patch~\cite{densecl, esvit}. Moreover, it echoes the design of Region of Interest (RoI) pooling~\cite{he2015spatial} in object detection, naturally facilitating the alignment of semantic types between patch- and image-level representations.

The key contribution of our method is the introduction of a novel objective that combines predictions from different levels to construct a multi-modal target distribution. As discussed earlier, a sharper target distribution in self-distillation leads to intensified compression. To mitigate this effect and augment the richness of representations, we model a natural image as an empirical distribution of visual features and learn an extra class token against the target distribution to produce multimodal predictions which reflects the semantic composition of an image. We show that this approach infuses "meaningful nuisances\footnote{\scriptsize {\it nuisance} indicates random info. of no direct interest in modeling but can be taken into account.}" of an image to its representation, hence enhancing the semantic richness. With the increased information flow through the bottleneck at class tokens, we coin the proposed method, \textbf{UDI} (/oo-d-ee/), conceptualized as \textbf{U}nsqueezed \textbf{Di}stillation-based SSL.

Extensive experiments show that UDI significantly improves versatility across both image-level and pixel-level tasks. Notably, UDI achieves competitive performance compared to existing state-of-the-art (SOTA) models on the ImageNet-1K (IN-1K)~\cite{imagenet}, achieving 77.6\% top-1 accuracy using a linear classifier and 75.6\% using a k-NN classifier. Furthermore, UDI shows advantages in low-shot learning, attaining 66.7\% accuracy using 1\% of the labels. On the downstream tasks, UDI not only excels in transfer learning but also demonstrate substantial improvement in dense prediction tasks, such as object detection, instance segmentation, and semantic segmentation. All experiments are performed with ViT due to its predominant performance in representation learning and multimodal integration. Specifically, we choose ViT-S/16 for its cost-effectiveness in terms of model size \emph{vs.} performance. Our findings are further illuminated via visual analysis and ablation studies elucidating the synergy between UDI and ViT.

\begin{figure}[tb]
  \centering
  \includegraphics[width=\textwidth]{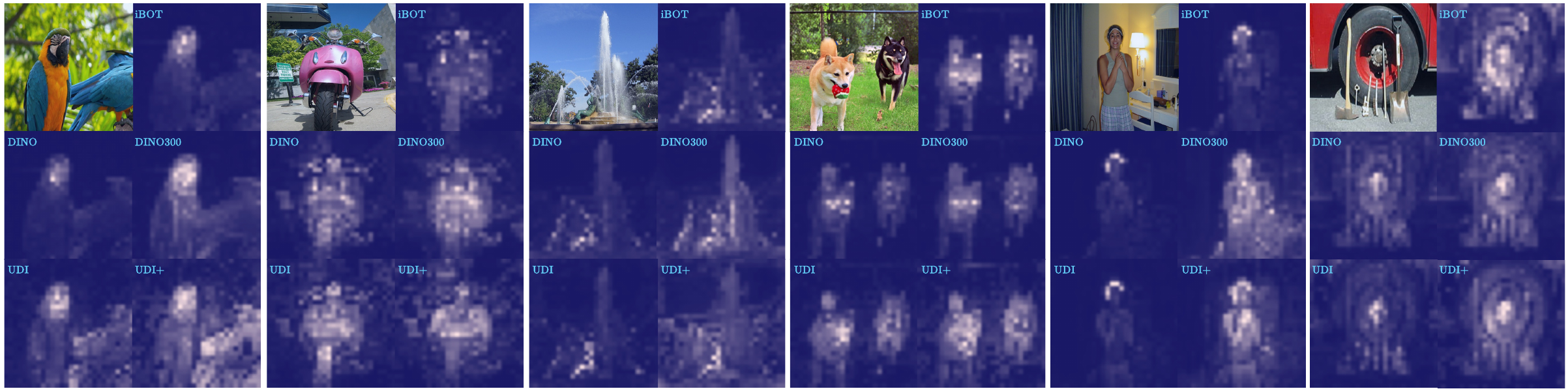}\vspace{-2mm}
  \caption{\footnotesize{\bf Visualization of \texttt{[cls]} attention} from the last layer of ViT-S/16 trained with self-distillation based SSLs. DINO300 denotes DINO trained with 300 epochs.}
  \label{fig:attention}
  \vspace{-5mm}
\end{figure}

\vspace{-3mm}
\section{Related Work}
\vspace{-2mm}
\noindent\textbf{SSL with Distillation.} Early works in this field utilize distillation in tandem with self-training~\cite{xie2020self} and SSL to propagate pseudo-labels or condensed knowledge from a pre-trained, fixed teacher model to a student model. Instead of using distillation as a post-processing step to SSL as in~\cite{fang2021seed,shen2021s2, chen2020big, noroozi2018boosting}, Caron et al.~\cite{dino} in DINO casts the knowledge distillation as a self-supervised objective by developing a teacher model online via dynamic updating. The teacher model is periodically refreshed using a moving average of the student model, and between updates, both models' predictions are optimized for consistency across augmented pairs of inputs. DINO's success in learning semantically meaningful representations is demonstrated by improved linear evaluation and downstream tasks, as well as visually coherent object-aligned attention. Subsequent works, such as iBOT~\cite{ibot}, DINOv2~\cite{oquab2023dinov2}, MST~\cite{mst} integrate Masked Image Modeling (MIM)~\cite{chen2020generativepixels, doersch2015unsupervised, pathak2016context} to factor the relevance and consistency of features at finer granularity into the objective, boosting the performance further. To capture meaningful representations at lower levels of granularity, several studies leverage multi-level clustering facilitated by self-distillation. This approach matches and groups pixels with similar semantics, as seen in EsViT~\cite{esvit}, DINO+SelfPatch\cite{selfpatch}, and CrOC~\cite{croc}, leading to improved results in dense prediction tasks. However, for all those works, beyond the intuitive summation of the objectives applied across different granularities, there is a lack of in-depth analysis of the synergistic interactions between those objectives and how effectively they contribute to improved performance. Due to page limit, a comprehensive review of related works is relegated to the Appendix \textcolor{red}{A}.

\vspace{-3mm}
\section{Methodology}
\vspace{-3mm}

\begin{figure}[tb]
  \centering
\includegraphics[width=\textwidth]{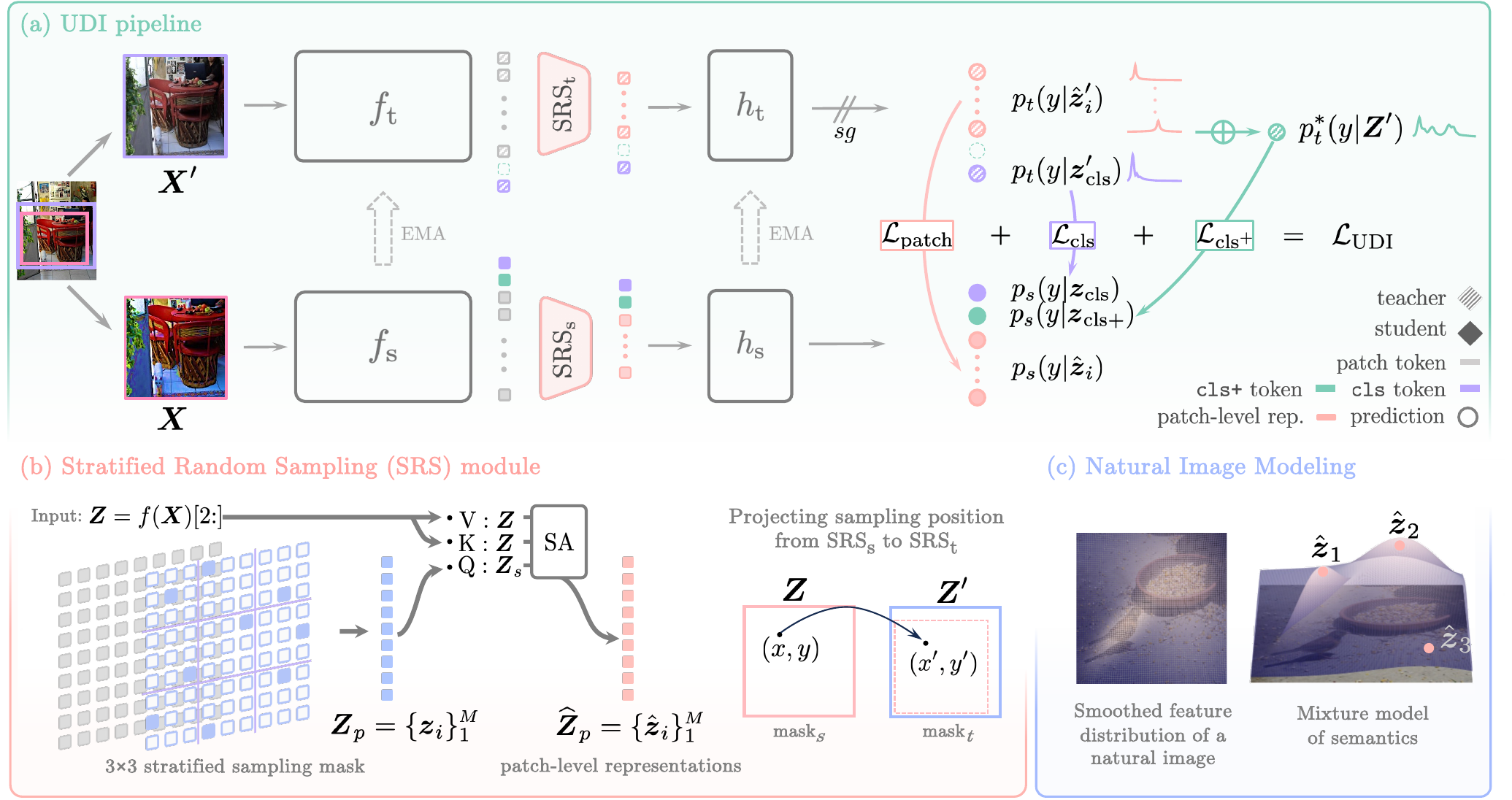}
\vspace{-4mm}\caption{\footnotesize {\bf UDI Framework}. UDI is an SSL method based on the joint-embedding strategy with multilevel self-distillation objectives. Specifically, for each image, UDI creates two views with one cropped out from the other, followed by two random augmentations, respectively, for student network $h_\text{s}\circ\text{SRS}_\text{s}\circ f_{\text{s}}$ and teacher network  $h_\text{t}\circ\text{SRS}_\text{t}\circ f_{\text{t}}$. UDI employs ViT with an extra class token $\bs{z}_{\texttt{cls}+}$ for encoder $f$. The dense features from $f$ are then sampled and processed by a Stratified Random Sampling (SRS) module to produce patch-level representations $\widehat{\bs{Z}}_p$. The class token $\bs{z}_{\texttt{cls}+}$ is learned to produce multimodal prediction against a target distribution $p^*_t({y}|\bs{Z}')$ constructed with patch-level predictions ${p}_t(y|\hat{\bs{z}}')$ and image-level prediction $p_t(y|\bs{z}'_{\text{cls}})$. The final UDI objective involves maximizing the agreement via cross-entropy loss between (i) $p_t(y|\bs{z}'_\text{cls})$, $p_s(y|\bs{z}_{\mathshift[0.5pt]{6}{6}{\text{cls}}})$, (ii) $p^*_t(y|\bs{Z}')$, $p_s(y|\bs{z}_{\text{cls}+})$, and (iii) ${p}_t(y|\hat{\bs{z}}'_i)$, ${p}_s(y|\hat{\bs{z}}_i)$, respectively.}
  \label{fig:udi_framework}
\vspace{-8mm}
\end{figure}

In this section, we first examine the limitations of current SSL approaches through the lens of information bottleneck, setting the stage for the development of UDI to guide the choice of a suitable learning principle. We then delve into the details of deriving context-aligned semantic constraints that guide the alignment of semantic types across different granularities for multi-level objective. This is followed by a novel objective designed to enhance representations with meaningful nuisance by promoting multimodal predictions. Finally, we present the complete learning framework of UDI, illustrated for clarity.

\vspace{-3mm}
\subsection{SSL through the Lens of Information Bottleneck (IB)}
\vspace{-2mm}
SSL with the joint embedding strategy falls in the framework of multiview information bottleneck (MIB) learning~\cite{ib}, which can be formulated as:

\noindent{\bf MIB:} Given two views $\{\bs{x}, \bs{x}'\}$ and their respective representations $\{\bs{y}, \bs{y}'\}$, the optimization problem under the \textit{mutliview assumption} aims to strike a balance between maximizing $I(\bs{x}'; \bs{y})$ the relevant information and minimizing $I(\bs{x}; \bs{y}|\bs{x}')$ the exclusive information $\bs{y}$ holding about $\bs{x}$ that is not inferable from $\bs{x}'$. Formerly, the objective to learn optimal representation $\bs{y}$ for $\bs{x}'$ can be approached with the Lagrangian relaxation as $\mathcal{L} = I(\bs{x};\bs{y}\mid \bs{x}') - \lambda I(\bs{x}';\bs{y})$,
and \emph{vice-versa} for $\mathcal{L}'$ to obtain $\bs{y}'$ for $\bs{x}$ due to symmetry. As shown in~\cite{federici2020learning}, the average of the two objectives $\{\mathcal{L}, \mathcal{L}'\}$ is upper-bounded by
\vspace{-3mm}
\begin{equation}
   \mathcal{ L}_{MIB} = -I(\bs{y};\bs{y}') + \beta D_{\text{SKL}}[P(\bs{y} \mid \bs{x})||P(\bs{y}' \mid \bs{x}')]
   \label{eq:mib}
   \vspace{-3mm}
   \footnote{\scriptsize In SSL with explicit clustering objectives, $\bs{y}$ corresponds to learned centroids, which we denote by their labels $y$ in later formulation.}
\end{equation}
with a re-parameterized Lagrange multiplier $\beta$ and the symmetric KL-divergence. 

\begin{figure}[tb]
\includegraphics[width=\textwidth]{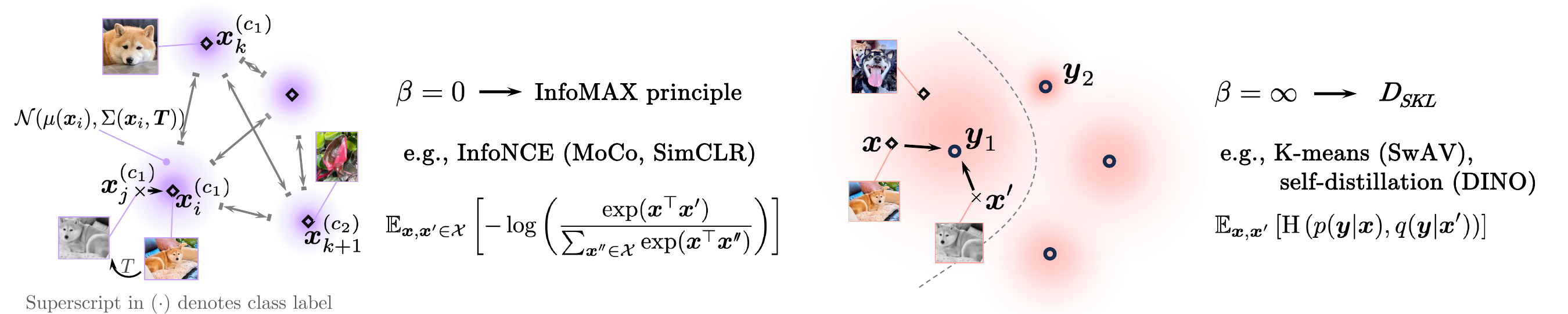}
    \caption{\footnotesize {\bf Clustering process of SSL objectives}. (\textcolor{mauve}{Left}) the InfoNCE objective drives an image $\bs{x}_i$ closer to its augmented views $\bs{x}_j$ (positive sample) while being far away from other images (negative samples). In the representation space, this leads to the formation of small clusters centered around each image $\bs{x}$, with their covariance depending on augmentation $\bs{T}$. (\textcolor{mauve}{Right}) explicit clustering-based SSL, such as DINO and SwAV, partitions the entire dataset into $K$ clusters.
    }\label{fig:ib_effect}
  \vspace{-7mm}
\end{figure}

Recent SSL methods tend to follow a reduced MIB principle, as shown in Fig.~\ref{fig:ib_effect}. When \(\beta\) is small (\(\beta\ll 1\)) or, equivalently, \(\lambda\) is large, the problem aligns with the InfoMax principle\cite{linsker1988self, oord2018representation}, which is characteristic of contrastive learning methods (e.g., MOCO~\cite{moco}, SimCLR~\cite{simclr}), and some non-contrastive learning methods~\cite{shwartz2023information} (e.g.,VICReg~\cite{vicreg}, SimSiam~\cite{simsiam}, BYOL~\cite{byol}). On the other hand, when \(\beta\) is large (or \(\lambda\) is small), Eq.~\ref{eq:mib} favors the compression of "redundant" information. This underpins SSL methods that utilize explicit clustering with Gaussian Mixture Model (GMM)~\cite{bishop2006pattern}, such as SwAV (K-means) and DINO (self-distillation). 

However, those objectives pose significant limitations. As demonstrated in~\cite{shwartz2023compress}, due to the inherent implicit clustering, InfoMax tends to retain irrelevant information, leading to overfitted clustering. 
Moreover, the success of InfoMax principle is more reliant on the inductive biases of the model and estimator than the training objectives themselves as highlighted in~\cite{tschannen2019mutual}. As for the explicit clustering based SSL objectives, they yield improved performance and generalizability upon InfoMax-based approaches since they learn more compressed and semantically meaningful representations. Nonetheless, these objectives are prone to over-compression. This arises particularly from an insufficient number of centroids and the use of sharpening technique that induces unimodal distribution~\cite{dino}, which restricts the richness in representations, compromising generalizability.

In light of the observations above, we propose a novel SSL framework that fosters a learning process balanced between information compression and nuisance preserving with semantic significance.

\vspace{-3mm}
\subsection{Multi-level SSL Objectives with Aligned Semantics}
\label{sec:aligned_semantics}
\vspace{-2mm}
Our method follows the explicit clustering IB principle due to its superiority of extracting meaningful information and its provision of greater control over clustering process. 
We adopt the DINO objective, reduced from the KL term of Eq.~\ref{eq:mib} with a large $\beta$, for representations at both the image level and patch level:
\vspace{-2mm}
\begin{equation}\label{eq:bi-level loss}
\begin{aligned}
    \mathcal{L}_{image} &=  \mathcal{L}_{\text{cls}} = \mathbb{E}_{\bs{z}_{\text{cls}},\bs{z}'_{\mathshift[0.5pt]{6}{6}{\text{cls}}}\in\mathcal{Z}_{\text{cls}}}\left[\text{H}(p_t(y|\bs{z}'_{\text{cls}}), p_s(y|\bs{z}_{\text{cls}}))\right],\\
    \mathcal{L}_{patch} &= \mathbb{E}_{\bs{z}_p,\bs{z}'_p\in\mathcal{Z}_p}\left[\text{H}(p_t(y|\bs{z}'_p), p_s(y|\bs{z}_p))\right],
\end{aligned}
\vspace{-2mm}
\end{equation}
where $\bs{z}_\text{cls}$ and $\bs{z}_p$ denotes the [\texttt{cls}]-token representation and patch-token representation, respectively, and  $\text{H}(a,b)=-a\log b$. 

\noindent\textbf{Lower-level Representations.} An image is essentially a distribution of features with clusters of patches representing distinct feature groups (as seen in semantic segmentation). Hence, it is more reasonable to learn meaningful representation per cluster than per patch.  With indefinite number of clusters, KDE-based algorithms, such as \emph{meanshift} (MS)~\cite{comaniciu2002mean, carreira2007gaussian}, are typically used to estimate the representations (modes) of these clusters. Given a patch $\bs{z}$ in image $\bs{Z}$, the one-step MS estimation towards the representation of $\bs{z}$'s pertaining cluster is given by 
\vspace{-3mm}
\begin{equation}\label{eq:ms}
\scaleto{{\rm MS}(\tau,\bm{z}_i,\bm{Z})\!=\! \sum_{j=1}^N\frac{e^{\sfrac{1}{\tau}\bm{z}_i^\top\mathbf{W}\bm{z}_j}}{\sum_{k=1}^N e^{\sfrac{1}{\tau}\bm{z}_i^\top\mathbf{W}\bm{z}_k}}\bm{z}_j = \bs{Z}\;\text{softmax}(\sfrac{1}{\tau}\bs{Z}^\top\mathbf{W}\bs{z}_i)}{30pt},
 \vspace{-3mm}
 \end{equation}
where $\bm{z}_i = \bs{Z}[:,i]$. The MS update rule above is essentially a special case of self-attention (SA)~\cite{attention, flsl}, which is defined as:
\vspace{-2mm}
\begin{equation}\label{eq: sa}
\hat{\bs{z}}_i\!=\text{SA}(\bs{z}_i,\bs{Z},\bs{Z}) = \!\mathbf{W}_V\bs{Z}\;{\rm softmax}\left(\sfrac{1}{\tau}\bs{Z}^{\top}\left(\mathbf{W}_K^\top \mathbf{W}_Q\right)\bs{z}_i\right).
\vspace{-2mm}
\end{equation}
Therefore, we can leverage an SA layer to obtain the cluster representation as the patch-level representation. Similar idea is explored in PixPro~\cite{pixpro}, where a similarity-based attention is leveraged to obtain smoothed representations that are more coherent with the underlying context. In SA, the map of attention weights is essentially a soft mask for representation pooling. This is also akin to a fine-grained average RoIPool in object detection~\cite{ren2015faster, he2017mask} to obtain regional representations for subsequent object classification or dense prediction tasks. As a result, the semantic types of these mask-pooled representations can naturally align with those at the image level, rendering our framework suitable for adopting a shared projector across different granularities. Another advantage is that the soft mask is a more reasonable semantic constraint. In contrast to the patch-wise constraints, such as the contrastive objective that differentiates each patch from the rest of an image as implemented in~\cite{densecl}, it enables adaptive semantic constraint that aligns with underlying context of an image. Consequently, both the {\bf aligned semantic types} of representations across different granularities and the {\bf aligned semantic constraints} with the context of image contribute to the improved performance as reported in Sec.~\ref{sec:experiments}.
\vspace{-3mm}
\subsection{Unsqueeze the \texttt{CLS} Token Bottleneck in Self-distillation }
\vspace{-2mm}
When a large number of centroids are used (e.g., $K\!=\!65,536$ for IN-1K), the primary source of compression in DINO is the skewed prediction sharpening between the teacher and student models, i.e., $0 < \tau_t < \tau_s \leq 1$. This design is crucial to avoid collapse. However, it also exacerbates the information compression beyond the clustering process by pursuing an increasingly sharper target, rendering the method susceptible to over-compression. To overcome this problem, we design a novel image-level objective to preserve meaningful nuisances alongside the image-level semantics by incorporating information from different granularities. 

\vspace{1mm}
\noindent{\bf Natural Image Modeling.} A natural image is a composition of semantics, \eg, objects and stuff\cite{coco}. Therefore, a \textit{comprehensive} representation of an image should reflect its semantic composition instead of focusing on the main characters. To achieve this, we approach the problem from the angle of multi-label classification to encourage a multimodal prediction for each image. Assuming that an image $\bs{Z}=f_{\theta}(\bs{X})\text{[1:]}$ is a composition of a set of semantic concepts $S_{\bs{Z}} = \{\Tilde{\bs{z}}_1, \dots, \Tilde{\bs{z}}_s\}$. Under the Markov constraint $y\leftrightarrow\Tilde{\bs{z}}\leftrightarrow\bs{Z}$, a compound posterior can then be constructed as
\vspace{-3mm}
\begin{equation}\label{eq:compound_posterior}
p(y|\bs{Z}) = \sum_{\mathshift[0.5pt]{8}{8}{\Tilde{\bs{z}}}}p(y, \Tilde{\bs{z}} | \bs{Z}) = \sum_{\mathshift[0.5pt]{8}{8}{\Tilde{\bs{z}}}} p(y|\Tilde{\bs{z}})p(\Tilde{\bs{z}}|\bs{Z}).
\vspace{-3mm}
\end{equation}   
However, it is challenging to define a complete set of semantics $S_{\bs{Z}}$ for each image to calculate $p(\Tilde{\bs{z}}|\bs{Z})$. We therefore approximate the above probability using the cluster representations $p(y|\hat{\bs{z}}_i)$ by treating an image as a mixture of semantic concepts, as illustrated in Fig.~\ref{fig:udi_framework}(c). By leveraging self-attention (SA) as the estimator of a cluster representation $\hat{\bs{z}}_i$ using patch $\bs{z}_i$ as query, Eq.~\ref{eq:compound_posterior} can then be approximated as
\vspace{-2mm}
\begin{equation}\label{eq:posterior approx}
p(y|\bs{Z}) \approx \frac{1}{M}\sum_{\mathshift[1pt]{7}{7}{i\!=\!1}}^{\mathshift[-2pt]{6}{6}{M}}p(y|\hat{\bs{z}}_i),\quad\text{where}\quad
\hat{\bs{z}}_i = \text{SA}(\bs{z}_i, \bs{Z}, \bs{Z}), \ \ i \sim \mathcal{U}_i.
\vspace{-3mm}
\end{equation}
Specifically, to improve efficiency, we randomly sample queries in $\bs{Z}$ instead of summing up and averaging over the entire image. To maintain the original composition of an image, the random sampling method should ensure that $p(\hat{\bs{z}}|\bs{Z}) \sim \frac{\text{area}(\hat{\bs{z}})}{\text{area}(\bs{Z})}$. Therefore, we use stratified random sampling, which is essentially uniform sampling within each grid (\eg, 3$\times$3) as denoted by $\mathcal{U}$ in Eq.~\ref{eq:posterior approx}. It effectively preserves the proportion of the semantic components of an image in the resulting representations while reducing aliasing effects, as detailed in~\cite{pharr2023physically}. 

We further incorporate the prediction $p(y|\bs{z}_{\text{cls}})$ from the regular class token to ensure that the image-level semantics are captured. With shared projector and aligned semantics as discussed in Sec.~\ref{sec:aligned_semantics}, we directly add them together using a blending factor $\alpha\in[0,1]$:
\vspace{-1mm}
\begin{equation}
    p^*({y}|\bs{Z}) = \alpha\;p({y}|\bs{Z}) + (1-\alpha)\;p(y|\bs{z}_{\text{cls}}).
    \vspace{-1mm}
\end{equation}
Finally, we arrive at the novel objective of UDI, and an extra class token, $\bs{z}_{\text{cls}+}$, is learned to produce multimodal predictions:
\vspace{-1mm}
\begin{equation}
    \mathcal{L}_{\text{cls}+} = \mathbb{E}_{\bs{Z}'\in\mathcal{Z},\bs{z}_{\text{cls}+}\in\mathcal{Z}_{\text{cls}+}}\left[\text{H}\left(p^*_t({y}|\bs{Z}'), p_s(y|\bs{z}_{\text{cls}+})\right)\right].
    \vspace{-1mm}
\end{equation}
Since the regular class token serves as a squeezed bottleneck structure between ViT and a projector, our method {\bf{unsqueezes}} it by implicitly adding more nodes and edges between dense features $\bs{Z}$ and the new class token in the last ViT layer, resulting in improved Forman curvature~\cite{topping2021understanding}.

It's worth noting that, to enrich the representation, a straightforward design derived from Eq.~\ref{eq:mib} is to directly combine the contrastive loss with explicit clustering loss to form a complete MIB objective (\eg, $\beta = 1$). However, the benefits of this formulation are limited due to InfoMAX's tendency to foster the retention of irrelevant information as data point identifier rather than nuisances of semantic significance~\cite{shwartz2023compress}, as reported in Appendix \textcolor{red}{E}.

\vspace{-3mm}
\subsection{The UDI Framework}
\vspace{-2mm}
\label{sec:framework}
As a result, with two class tokens $\{\bs{z}_{\text{cls}}, \bs{z}_{\text{cls}+}\}$, the image-level loss is defined as:
\vspace{-2mm}
\begin{equation}
    \begin{aligned}
   \mathcal{L}_{image} &= \mathcal{L}_{\text{cls}} + \mathcal{L}_{\text{cls}+}\\
   &\hspace{-9mm}=\scaleto{
\mathbb{E}_{\bs{z}_\text{cls},\bs{z}'_\text{cls}\in\mathcal{Z}_\text{cls}}\left[\text{H}(p_t(y|\bs{z}'_\text{cls}), p_s(y|\bs{z}_\text{cls}))\right] + \mathbb{E}_{\bs{Z}'\in\mathcal{Z},\bs{z}_{\text{cls}+}\in\mathcal{Z}_{\text{cls}+}}\left[\text{H}\left(p^*_t({y}|\bs{Z}'), p_s(y|\bs{z}_{\text{cls}+})\right)\right].}{10pt}
   \end{aligned}
   \vspace{-1mm}
\end{equation}
Meanwhile, the patch-level loss is applied to the patches that are randomly sampled for constructing $p_t(y|\bs{Z}')$:
\vspace{-2mm}
\begin{equation}
    \mathcal{L}_{patch} = \mathbb{E}_{\bs{z}\in\bs{Z},\bs{Z}\in\mathcal{Z}}\left[\text{H}(p_t(y|\hat{\bs{z}}\mathshift[-0.5pt]{8}{8}{'}), p_s(y|\hat{\bs{z}})\right],
    \vspace{-1mm}
\end{equation}
where the position of query $\bs{z}'$ for the teacher model is projected from the position of $\bs{z}$ for the student model. Finally, the total loss of UDI is given by:
\vspace{-1mm}
\begin{equation}
    \mathcal{L}_{UDI} = \mathcal{L}_{image} + \mathcal{L}_{patch} = \mathcal{L}_{\text{cls}} + \mathcal{L}_{\text{cls}+} + \mathcal{L}_{patch}.
    \vspace{-1mm}
\end{equation}

The UDI framework, as illustrated in Fig.~\ref{fig:udi_framework}, adopts the joint-embedding architecture composed of a teacher model $f_t$ and a student model $f_s$; both are vision transformers with an extra class token $\bs{z}_{\text{cls}+}$. At each iteration, two models encode two views $\{\bs{X}', \bs{X}\}$ of the input image, respectively. The resulting dense features $\bs{Z}'= f_t(\bs{X}')[2\text{:}]$, $\bs{Z}= f_s(\bs{X})[2\text{:}]$ are then selected and processed by the Stratified Random Sampling (SRS) module to extract patch-level representations $\{\hat{\bs{z}}'\}$, $\{\hat{\bs{z}}\}$. Note that patches of $f_t$ are selected by their positions projected from that of the randomly sampled patches from $f_s$. Subsequently, image-level and patch-level representations from both models are passed to their respective projectors $h_t$, $h_s$ to produce prediction over $K$ centroids. Projectors are shared between two representation levels. The predictions from teacher on image level $p_t(y|\bs{z}_{\text{cls}})$ and patch level $\{p_t(y|\hat{\bs{z}}')\}$ are then combined to form the target distribution $p_t^*(y|\bs{Z}')$. The UDI loss comprises three cross-entropy losses respectively for image-level predictions $\{p_t(y|\bs{z}'_{\text{cls}}), p_s(y|\bs{z}_{\text{cls}})\}$, patch-level predictions $\{p_t(y|\hat{\bs{z}}'), p_s(y|\hat{\bs{z}})\}$, and multimodal predictions $\{p_t^*(y|\bs{Z}'), p_s(y|\bs{z}_{\text{cls}+})\}$. All the parameters from the teacher branch are iteratively updated from the student through EMA. The pseudo-code of UDI is provided in Appendix \textcolor{red}{B}.

\vspace{-3mm}
\section{Experiments}\label{sec:experiments}
\vspace{-2mm}
We present the performance evaluation of UDI on the standard benchmarks for classification, transfer learning, object detection and segmentation, and video segmentation with comparison against several representative state-of-the-art (SOTA) SSL approaches. We also include major ablation study and analysis to elucidate the success of UDI framework.

\begin{table*}[t!]
\setlength{\arrayrulewidth}{.01pt}
\begin{minipage}[c]{0.46\textwidth}
\caption{\footnotesize Linear probing and $k$-NN evaluation on ImageNet-1K.\vspace{-4mm}}
\resizebox{\textwidth}{!}{\begin{tabular}{c|l|c|cccc}
    \toprule[1.pt]
    \multirow{2}{*}{~} & Method &\scalebox{0.8}{Arch.} & \#Views & Epoch$^{\dag}$ &Linear  &$k$-NN \\
    \cline{2-7}
    ~ & Supervised  &\multirow{21}{*}{\rotatebox[origin=c]{90}{\makebox[0.2\columnwidth]{\scalebox{0.8}{ViT-S/16}}}} & 1 & 300 & 79.3 & 79.3 \\
    ~ & BEiT & & 1 & 800 &24.2 &6.9 \\
            \cline{1-2}\cline{4-7}
\multirow{7}{*}{\rotatebox[origin=c]{90}{\makebox[0.25\columnwidth]{\scalebox{0.9}{Implicit Clustering}}}}             & SimCLR~\cite{simclr} & & 2 & 600 & 69.0 & --- \\
	~ & MoBY~\cite{moby}  & & 2 & 600 & 72.8 & --- \\
 	~ & BYOL~\cite{byol}  & & 2 & 600 & 71.0 & --- \\
	~ & BYOL~\cite{byol}  & & 2 & 2000 & 71.4 & 66.6 \\
	~ & MoCov2~\cite{mocov2}  & & 2 & 1600 & 72.7 & 64.4 \\
	~ & MoCov3~\cite{mocov3} & & 2 & 600 & 72.5 & --- \\
        ~ & MoCov3~\cite{mocov3} & & 2 & 1200 & 73.4 & --- \\
	\cline{1-2}\cline{4-7}
\multirow{10}{*}{\rotatebox[origin=c]{90}{\makebox[0.25\columnwidth]{\scalebox{1}{Explicit Clustering}}}} 
	   & SwAV~\cite{swav}  &  & 8 & 2400 & 73.5 & 66.3 \\
    ~ &\cellcolor{lgray}DINO~\cite{dino} & &\cellcolor{lgray}2 & \cellcolor{lgray}600 & \cellcolor{lgray}72.5 & \cellcolor{lgray}--- \\
    ~ &\cellcolor{lgray}DINO~\cite{dino} & &\cellcolor{lgray}12 & \cellcolor{lgray}400 & \cellcolor{lgray}74.6 & \cellcolor{lgray}--- \\
	~ &\cellcolor{lgray}DINO~\cite{dino} & &\cellcolor{lgray}12 & \cellcolor{lgray}1200 & \cellcolor{lgray}76.1 & \cellcolor{lgray}72.8 \\
        ~ &\cellcolor{lgray}DINO+reg. & &\cellcolor{lgray}12 & \cellcolor{lgray}1200 & \cellcolor{lgray}76.1 & \cellcolor{lgray}72.9 \\
	~ & DINO~\cite{dino} &  & 12 & 3200 & 77.0 & 74.5 \\
        ~ & MST~\cite{mst} & & 12 & 1200 & 76.3 & 75.0 \\
        ~ & iBOT~\cite{ibot} &  & 2 & 1600 & 76.2 & 72.4 \\
	~ & iBOT~\cite{ibot} &  & 12 & 1200 & 77.4 & 74.6 \\
	~ & iBOT~\cite{ibot} &  & 12 & 3200 & \bf{77.9} & 75.2 \\
 	~ & MSN~\cite{msn} &  & 11 & 900 & 76.9 & - \\
        ~ & \cellcolor{flesh}{\bf UDI$_+$}  &  & \cellcolor{flesh}12 & \cellcolor{flesh}1200 & \cellcolor{flesh}75.1 &  \cellcolor{flesh}68.9 \\
	~ & \cellcolor{flesh}{\bf UDI}  &  & \cellcolor{flesh}12 &\cellcolor{flesh}1200 & \cellcolor{flesh}77.6 & \cellcolor{flesh}\bf{75.6} \\
	\bottomrule[1.pt]
    \end{tabular}}
    \scriptsize{Epoch$^{\dag}$ denotes effective pre-training epochs calculated following~\cite{ibot}. Evaluation results using the regular class token $\bs{z}_{\text{cls}}$ and extra class token $\bs{z}_{\text{cls}+}$ are denoted by UDI and UDI$_+$, respectively. "+reg." indicates pretrain with a register token.}
    \label{tab:linear}
\end{minipage}
\vspace{-4mm}
\begin{minipage}[r]{0.51\textwidth}
\captionsetup{font=footnotesize}
\caption{\footnotesize Low-shot evaluation on IN-1K.}
\vspace{-4mm}
\label{table:low-shot}
\resizebox{\textwidth}{!}{
\centering
    \begin{tabular}{l | c | c c | cc}
					\toprule[1.pt]
					\multirow{2}{*}{\textbf{Method}}  & \multirow{2}{*}{\textbf{Arch.}} &   \multicolumn{2}{c|}{\textbf{logistic regression}} &    \multicolumn{2}{c}{\textbf{fine-tuning}} \\
					
					& & 1\% & 10\% &    1\% & 10\%  \\
					\midrule 
					SimCLRv2~\cite{simclrv2} & RN50 & --- & --- & 57.9 & 68.1  \\
					BYOL~\cite{byol} & RN50 & ---  & ---  & 53.2 & 68.8 \\
					SwAV~\cite{swav} & RN50 &---  & ---  & 53.9 & 70.2 \\
					SCLRv2+SD & RN50 & --- & --- & 60.0 & 70.5 \\
					DINO~\cite{dino}  & ViT-S/16 &   64.5 & 72.2  &    60.3 & 74.3\\
					iBOT~\cite{ibot}    & ViT-S/16 &   65.9 & 73.4  &    61.9 & 75.1  \\
                    MSN~\cite{msn}    & ViT-S/16 &   \textbf{67.2} & ---  &  --- & ---  \\
					\rowcolor{flesh}\textbf{UDI}  & ViT-S/16   &   66.7 & \textbf{74.1} &  \textbf{65.8} & \textbf{76.4}\\
                    \rowcolor{flesh}\textbf{UDI$_+$}  & ViT-S/16   &   66.1 & 73.8 &  65.2 & 76.2\\
					\bottomrule[1.pt]
				\end{tabular}}
\vspace{3pt}
\caption{Transfer learning (classif. acc. \%)}\label{table:transfer}
\vspace{-4mm}
\resizebox{\textwidth}{!}{
		\begin{tabular}{l|cccccc}
					\toprule[1.pt]
					\multirow{2}{*}{\textbf{Method}} &  \multicolumn{6}{c}{\textbf{ViT-S/16}} \\ 
					& {{Cif$_{10}$}} & {{Cif$_{100}$}} & {{INat$_{18}$}} & {{INat$_{19}$}} & {{Flowers}} & {{Car}}\\ 
					\hline
					Supervised~\cite{dino}      & 99.0 & 89.5 & 70.7 & 76.6 & 98.2 & 92.1 \\ 
					BEiT~\cite{beit}       & 98.6 & 87.4 & 68.5 & 76.5 & 96.4 & 92.1 \\
					DINO~\cite{dino}       & 99.0 & 90.5 & 72.0 & 78.2 & 98.5 & 93.1 \\
                        DINO+reg.       & 98.8 & 90.5 & 72.1 & 78.2 & 98.5 & 93.2 \\
					iBOT~\cite{ibot}       & {\bf 99.1} & 90.7 & 73.7 & 78.5 & 98.6 & {94.0}\\
					\rowcolor{flesh}
					\textbf{UDI}    & {\bf 99.1} &  {90.8} & {74.1}  & {78.9} &{98.6} &  {\bf 94.1} \\
                     \rowcolor{flesh}\textbf{UDI}$_+$ & {\bf 99.1} &{\bf 91.3} &{\bf 74.8} &{\bf 79.7} &{\bf 98.9} &94.0\\
					\bottomrule[1.pt]
		\end{tabular}
  }
\vspace{3pt}
\caption{Fine-tuning on IN-1K.}\label{table:ft}
\vspace{-4mm}
\resizebox{\textwidth}{!}{
\begin{tabular}{c|c|c|cc|cc}
\toprule[1.pt]
                            \multirow{2}{*}{~}&
                            Method& 
                            Arch.&
                            Epo.$^{\dag}$&
                            Acc.$(\%)$&
                            Epo.$^{\dag}$&
                            Acc.$(\%)$\\
                            \hline
Supervised & --- &\multirow{5}{*}{\rotatebox[origin=c]{90}{\makebox[0.2\columnwidth]{\scalebox{0.8}{ViT-S/16}}}} & ---  &79.9 &--- & ---\\ \cline{1-2}\cline{4-7}
\multirow{4}{*}{Clustering}
&DINO\cite{dino}& &1200 & 81.7 &3200 &82.0\\
&iBOT\cite{ibot}& &1200 & 82.0 &3200 &82.3\\
&\cellcolor{flesh}{\bf UDI}& &\cellcolor{flesh} 1200 &\cellcolor{flesh} 82.6 &800 &82.2\\ 
&\cellcolor{flesh}{\bf UDI$_+$}& &\cellcolor{flesh} 1200 &\cellcolor{flesh} {\bf 83.0} &800 &82.7\\
\bottomrule[1.pt]
\end{tabular}}
\end{minipage}
\vspace{-2mm}
\end{table*}

\vspace{-4mm}
\subsection{Experiment Setup}
\vspace{-2mm}
\noindent\textbf{Architectures.} We employ ViT-S/16 to assess the effectiveness of UDI due to its cost-efficiency balance between training cost and performance. Since an additional class token $\bs{z}_{\text{cls}+}$ is learned alongside the regular one $\bs{z}_{\text{cls}}$, we utilize both class tokens for evaluation on image-level downstream tasks. The SRS module consists of a standard multi-headed self-attention layer without the skip connection. Projectors follow the implementation of DINO with K = 65,536 and are shared for both image-level and patch-level representations.

\vspace{3pt}
\noindent\textbf{Pretraining Setting.} We use ImageNet-1K for UDI pretext training. Following DINO, we pretrain ViT-S/16 for 300 epochs using the AdamW optimizer with a momentum of 0.9, a cosine-scheduled \cite{sgdr} weight decay from 0.04 to 0.1, and a cosine-scheduled learning rate with 10 epochs of linear warm-up to the base value determined by $8\times10^{-4}\times\text{batch-size}/256$. The data augmentation strategy follows BYOL\cite{byol}. Specifically, given an image as input, a global view for teacher model is first created via random crop, upon which, another global view and multiple local views for student model are randomly cropped. Global views and local views are resized to $224 \times 224$ and $96\times96$, respectively, with a random horizontal flip, followed by color distortion, Gaussian-blur, and a solarization operation. We create 10 local crops, which add up to 12 views in total, resulting in $r = 2 + (\frac{96}{224})^2 \times 10 = 3.84 \approx 4$ effective views calculated following iBOT.

For UDI hyperparameters, we set $\tau_s=0.1$ and the final $\tau_t = 0.07$ with a linear warm-up from $0.04$ for both image-level and patch-level predictions. We adopt centering with a momentum of 0.9 following DINO. The blending factor $\alpha$ is set to $0.5$ for constructing the multimodal target distribution. SRS module uses $3\times 3$ sampling grid. UDI has a similar training cost to iBOT, as the projector/prediction heads and the self-attention layer in the SRS module are much smaller than the backbone. Complete implementation details are provided in Appendix \textcolor{red}{C}. 

\vspace{-0.9em}
\subsection{Evaluation on ImageNet-1K}\label{resultsonImagenet}
\vspace{-0.4em}
We compare UDI with SOTA methods on ImageNet-1K under three classification settings: linear probing, $k$-NN, low-shot learning, and fine-tuning. Unless specified, all the image-level tasks are evaluated using the regular class token.

\vspace{5pt}
\noindent\textbf{Linear Probing.} This involves training a linear classifier with frozen features generated by the backbone for 300 epochs on ImageNet-1K. For a fair comparison, we follow the conventional settings in DINO and iBOT in our experiments. Table~\ref{tab:linear} reports the top-1 test accuracy on ImageNet-1K. We evaluate our method for both the regular class token, denoted by UDI, and the multimodal-promoted class token, denoted by UDI$_{+}$. With the same training cost, UDI consistently outperforms other methods. Specifically, UDI achieves 77.6\% top-1 accuracy with 1,200 effective epochs. Whereas UDI$_+$ falls short in linear evaluation due to its enriched multimodal representations, leading to lower linear separability. We also provide results of other SOTAs with a larger number of effective epochs, e.g., 3,200. Notably, even though our model is only trained for 1,200 epochs, the accuracy of UDI is very competitive to methods trained for 3,200 epochs.

\vspace{5pt}
\noindent\textbf{{\it k}-NN Evaluation.} We report $k$-NN evaluation, another widely used evaluation protocol in SSL, to test the quality of representations learnt by UDI. For a fair comparison, we follow DINO to sweep over different numbers, \ie, $10$, $20$, $50$, $100$, of nearest neighbors for UDI-learned representations. From the last column of Table~\ref{tab:linear}, we can see that UDI achieves the highest accuracy of 75.6\%, outperforming the 1st and 2nd runner-ups, iBOT and MST, by 0.4\% and 0.6\% respectively, with the same epochs. As discussed in Sec.~\ref{sec:ablation}, the aligned multi-level semantics and multi-model prediction pretext facilitate the disentanglement between image-level semantics and lower-level significant nuisances, leading to the improved $k$-NN performance. Note that UDI$_+$ incurs the compromised performance for the same reason as discussed in the linear probing evaluation.

\noindent\textbf{Low-shot Learning.} Table~\ref{table:low-shot} reports evaluation on ImageNet-1K using 1\% and 10\% of the labels, which are standard benchmarks to evaluate label efficiency in SSL. We follow the implementation in DINO and iBOT to fine-tune a linear head on top of a pretrained encoder, which can be either frozen or end-to-end trainable, and then evaluating the fine-tuned model on the test data. The results demonstrate that for both 1\% and 10\% training data, UDI consistently outperforms most previous SOTA methods under both settings. Notably, under the fine-tuning setting with 1\% labeled data, UDI achieves a remarkable improvement of 3.9\% in accuracy over iBOT, showcasing the effectiveness of UDI in low-shot learning scenarios. 

\noindent\textbf{Fine-tuning} 
To further evaluate the quality of UDI-pretrained model, we  follow the fine-tuning protocol used in BEiT~\cite{beit} and iBOT. The protocol uses a layer-wise learning rate decay, weight decay and AdamW optimizer to train ViT on ImageNet-1K. By default, we use a layer-wise decay of 0.75 with a training epoch of 200 for ViT-S/16. Table~\ref{table:ft} summarizes performance in classification accuracy, where ``Supervised" indicates supervised training from randomly initialized parameters using 300 epochs with its result quoted from DeiT~\cite{deit}. ViT-S achieves new SOTA results of 83.0\%, improving over iBOT  by  0.7\%. Even with fewer pretraining epochs, UDI achieves a 0.4\% improvement compared to the prior SOTA. This suggests UDI's effectiveness in preserving in-distribution information. 

\vspace{-4mm}
\subsection{Downstream Tasks}
\vspace{-0.5em}
We evaluate the performance of UDI on various downstream tasks, including transfer learning and dense prediction tasks. The details of training and implementation can be found from Appendix \textcolor{red}{C}. 

\noindent\textbf{Transfer Learning.} We examine the transferability of the model pretrained with UDI on ImageNet-1K to various smaller datasets. We fine-tune and evaluate the model by following the training recipe and protocol used in Dosovitskiy et al.~\cite{vit}. Table~\ref{table:transfer} reports the results, indicating that UDI consistently outperforms other SSL approaches, achieving state-of-the-art performance with reduced pretraining time. While performances on CIFAR-10/-100 and Flowers have plateaued, UDI$_+$ demonstrates a more substantial performance gain, particularly on the more challenging datasets, such as iNaturalist 18 and 19. This indicates that the pretrained representations are enhanced with meaningful nuisances. This enhancement can be attributed to the learning of multimodal prediction, which facilitates effective exploitation of more useful information in an image with the pretrained model.

\vspace{-1.4em}
\subsubsection{Object Detection \& Instance  Segmentation.}  
We assess the performance of the UDI-pretrained model in object detection and instance segmentation tasks on the COCO dataset~\cite{coco}. These tasks require the ability to accurately locate and distinguish objects. We adopt the \textbf{Mask R-CNN}~\cite{maskrcnn} detection framework with a ViT-S/16 backbone for our evaluation, following the implementation in \cite{selfpatch}. As shown in Table~\ref{tab:coco_det_seg}, UDI consistently outperforms all the representative SSL methods with or without considering multi-level objectives, by large margins. Specifically, UDI significantly improves the performance over the SSL methods with image-level objectives, \eg, MSN, by +1.5\%. Under similar pretraining time, UDI consistently outperforms SelfPatch (utilizing bi-level objectives) by +1.1\% and 0.9\% points on detection (AP$^{bb}$) and instance segmentation (AP$^{mk}$) tasks, respectively. As demonstrated in ablation study, the improvement of UDI largely comes from the reasonable semantic constraints enabled by self-attention which operates as the soft-masked pooling, similar to the RoI pooling operations essential for object detection. The complete comparison across different backbones of similar sizes are provided in Appendix \textcolor{red}{D}.

\begin{table*}[tb!]
\begin{minipage}[l]{0.49\textwidth}
    \caption{\footnotesize Object detection and instance segmentation on MS-COCO with Mask R-CNN framework under 1x schedule.}
    \vspace{-8pt}
    \footnotesize
    \resizebox{\textwidth}{!}{\begin{tabular}{l|l|cc|cc}
    \toprule[1.pt]
        ~ & {Method} & {Backbone} & {\#Epo.} & AP$^{bb}$  & AP$^{mk}$\\
        \hline
        \multirow{6}{*}{\rotatebox[origin=c]{90}{\makebox[0.25\columnwidth]{\scalebox{0.9}{Image-level SSLs}}}} & Moco-V2~\cite{mocov2} & RN50 & 200  & 38.9  & 35.5  \\
        & SwAV~\cite{swav} & RN50 & 200 & 38.5  & 35.4  \\
        & MoCov3~\cite{mocov3} & ViT-S/16 & 300 & 39.8& 37.1  \\
        & MoBY~\cite{moby} & ViT-S/16 & 300 & 41.1  & 37.3  \\
        & DINO~\cite{dino} & ViT-S/16 & 800 & 40.8 & 37.4   \\
        & MSN~\cite{msn} & ViT-S/16 & 800 &41.7 &38.1 \\	
        \hline
        \multirow{6}{*}{\rotatebox[origin=c]{90}{\makebox[0.25\columnwidth]{\scalebox{0.9}{Multi-level SSLs}}}} & DenseCL~\cite{densecl} & RN50 & 200  & 40.3 & 36.4 \\
        & ReSim~\cite{resim} & RN50 & 200  & 40.3  & 36.4  \\
        & DetCo~\cite{detco} & RN50 & 200 & 40.1 & 36.4  \\
        & iBOT~\cite{ibot} & ViT-S/16 & 800 &41.7 &38.1 \\
        & SelfPatch~\cite{selfpatch} & ViT-S/16 & 200 & 42.1 & 38.5  \\
        & \cellcolor{flesh}UDI & \cellcolor{flesh}ViT-S/16 & \cellcolor{flesh}300 & \cellcolor{flesh}{\bf 43.2}  &\cellcolor{flesh}{\bf 39.4}  \\	
        \bottomrule[1.pt]
    \end{tabular}}
    \\
    \scriptsize{\# Epo. refers to the number of pretraining epochs on ImageNet-1K.}
    \label{tab:coco_det_seg}
\end{minipage}
\begin{minipage}[l]{0.49\textwidth}
\caption{\footnotesize ADE20K semantic segmentation with SSL pre-trained models on ImageNet.}
\label{tab:ade_seg}
    \vspace{-7pt}
\resizebox{\textwidth}{!}{\begin{tabular}{l|cc|ccc}
        \toprule[1.pt]
        Method & Arch. & Backbone. & mIoU & aAcc & mAcc \\
        \hline
        MoCo-v2~\cite{mocov2} & FPN & RN50  & 35.8 & 77.6 & 45.1 \\
        SwAV~\cite{swav} & FPN & RN50  & 35.4 & 77.5 & 44.9 \\
        DenseCL~\cite{densecl} & FPN & RN50  & 37.2 & 78.5 & 47.1 \\
        MocoV3~\cite{mocov3} & FPN & ViT-S/16  & 35.3 & 78.9 & 45.9 \\
        MoBY~\cite{moby} & FPN & ViT-S/16  & 39.5 & 79.9 & 50.5 \\
        DINO~\cite{dino} & FPN & ViT-S/16  & 38.3 & 79.0 & 49.4 \\
        DINO~\cite{dino} & UN & ViT-S/16 & 42.3 & 80.4 & 52.7 \\
        iBOT~\cite{ibot} & UN & ViT-S/16 & 42.9  & 81.1  &  53.4 \\
        SelfPatch~\cite{selfpatch} & FPN & ViT-S/16  & 41.2 & 80.7 & 52.1 \\
        SelfPatch~\cite{selfpatch} & UN & ViT-S/16 & 43.2 & 81.5 & 53.9 \\
        CrIBo~\cite{selfpatch} & FPN & ViT-S/16 & 42.6 & 80.8 & 52.9 \\
        \rowcolor{flesh}UDI & FPN & ViT-S/16  &42.6 & 81.9 & 53.4  \\
        \rowcolor{flesh}UDI & UN & ViT-S/16  &{\bf 43.7}  &{\bf 82.6}  & {\bf 54.5}  \\
        \bottomrule[1.pt]
    \end{tabular}}
 \vspace{-1pt}\\
 \scriptsize{Models with FPN and UperNet (UN) framework are trained for 40K and 160K iteration, respectively.}
\end{minipage}
\vspace{-6mm}
\end{table*}

\vspace{-1pt}
\noindent\textbf{Semantic Segmentation on ADE20K.} To evaluate the semantic segmentation performances of UDI pre-trained models, we follow the standard benchmark, ADE20K~\cite{ade20k}, and fine-tune the pretrained model in different frameworks (FPN~\cite{fpn} and UperNet~\cite{upernet}) with 40k and 160k iterations, respectively. Results are reported in Table~\ref{tab:ade_seg} in terms of three metrics: the mean intersection over union (mIoU), all pixel accuracy (aAcc), and mean class accuracy (mAcc). Still, UDI continually outperforms SOTAs especially when using the lighter prediction head FPN. Specifically, UDI achieves +4.3\% over DINO and +1.4\% over SelfPatch with FPN. With less impact from the prediction framework, this suggests the improved feature quality. Similarly, we attribute this improvement to the context-aligned semantic constraint as demonstrated in Fig.~\ref{fig:semantic_constraint}.

\vspace{-1pt}
\noindent\textbf{Video Object Segmentation.} We extend our evaluation to DAVIS-2017~\cite{davis} to assess the transferability of learned features in capturing semantics with consistency across frames via nearest neighbor retrieval. Following the protocol in Jabri et al.~\cite{jabri2020space}, we report the results in Table~\ref{table: davis} in terms of two metrics: mean region similarity $\mc{J}_m$ and mean contour-based accuracy $\mc{F}_m$. Consistently, UDI shows better feature transferability than DINO and iBOT, indicating the UDI-pretrained encoder captures consistent and meaningful semantics at finer granularities, \ie, the nuisances of semantic significance.

\vspace{1mm}
\noindent In Fig.~\ref{fig:apacc}, we visualize via bubble plot the performances of different SSL-pretrained models on both image-level and lower-level (dense prediction) downstream tasks, represented by linear probing and object detection, respectively. UDI achieves more balanced performance between the two tasks, suggesting its effectiveness in capturing more comprehensive information from images. 
\begin{table*}[t]
\begin{minipage}[l]{0.47\textwidth}
\caption{\footnotesize DAVIS 2017 Video instance Segmentation.}
\label{table: davis}
\vspace{-4mm}
\resizebox{\textwidth}{!}{
\begin{tabular}{l|cccc}
\toprule[1.pt]
                    Pretrain 
                            &Arch.
                            &$(\mathcal{J}\&\mathcal{F})_m$
                            &\;$\mathcal{J}_m$\;\;
                            &\;\;$\mathcal{F}_m$\\
                            \hline
IN-1K MAE\cite{mae}   &ViT-S/16 &50.8 &49.1 &52.5\\
IN-1K MoCo-v3\cite{mae}   &ViT-S/16 &53.5 &51.2 &55.9\\
IN-1K MoBY\cite{moby}   &ViT-S/16 &54.7 &52.0 &57.3\\
IN-1K DINO\cite{dino}   &ViT-S/16 &61.8 &60.2 &63.4\\
IN-1K iBOT\cite{ibot} &ViT-S/16 &62.1 &60.9 &63.3\\
IN-1K SelfPatch\cite{selfpatch} &ViT-S/16 &62.7 &60.7 &64.7\\
\rowcolor{flesh} IN-1K UDI   &ViT-S/16 &\textbf{62.9} &\textbf{61.1} &\textbf{64.9}\\
\bottomrule[1.pt]
\end{tabular}}
\vspace{-1pt}\\
\scriptsize{Metrics: mean region similarity $\mc{J}_m$ and mean contour-based accuracy $\mc{F}_m$.}
\end{minipage}
\begin{minipage}[r]{0.47\textwidth}
\includegraphics[width=\textwidth]{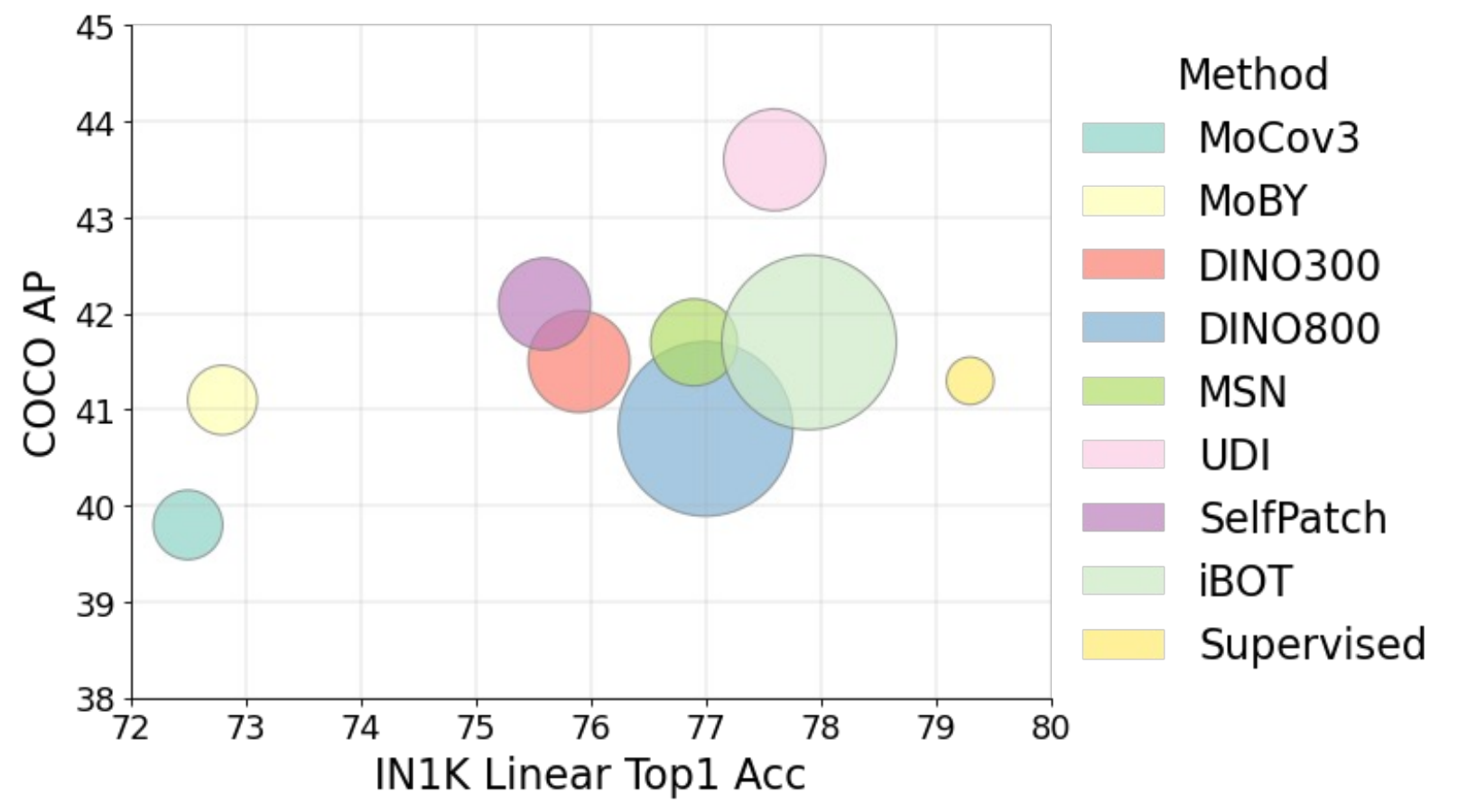}
  \captionof{figure}{\footnotesize Linear probing and object detection performance of SSL-pretrained models, with bubble size $\propto$ \#epochs in SSL.}
  \label{fig:apacc}
\end{minipage}
\vspace{-18pt}
\end{table*}

\vspace{-3mm}
\subsection{Ablation Study}\label{sec:ablation}
We investigate the effect of the novel components and their contributions in UDI: (i) the impact of the proposed semantic constraint, (ii) the impact of shared projector (for aligning the semantic types) in multi-level objective, and (iii) the impact of multimodal objective. We also include the study of the sensitivity of the blending factor $\alpha$. In all the ablations, we pretrain ViT-S/16 for 100 epochs using UDI with multi-crop setting and report linear evaluation on ImageNet-1K and object detection on COCO with 1$\times$ schedule. More ablation results are provided in the Appendix \textcolor{red}{E}\footnote{\scriptsize Additionally, we evaluate the "InfoNCE + DINO" objective directly constructed from Eq.~\ref{eq:mib} to provide insights into its inherent limitations and contrast with our method.}. 

\vspace{1mm}
\noindent{\bf Semantic Constraint \& Shared Projector.} We jointly examine the effects of semantic constraints and the shared projector. Table~\ref{table: semantics} presents a comparison of different types of patch-level semantic constraints employed by existing SSL methods, mainly categorized into similarity-based (EsViT, SelfPatch) and position-based (PixContrast~\cite{pixpro}) constraints. For a fair comparison, we employ the DINO loss for both image and patch levels and train for 100 epochs for all the experiments, reporting ImageNet-1K linear evaluation and COCO AP under a 1$\times$ schedule. In PixContrast, we implement the positive token as the average pooling of tokens within the projected vicinity. Notably, UDI secures the most significant performance improvement over the vanilla DINO, especially in dense prediction tasks. Unlike PixContrast, UDI employs soft-masked pooling via self-attention, resulting in a semantic constraint that is more aligned with the underlying context, as depicted in Fig.~\ref{fig:semantic_constraint}. The performance is further enhanced with a shared projector for both image-level and patch-level representations.

\vspace{1mm}
\noindent{\bf Multimodal Objective $\mc{L}_{\rm cls+}$.} 
The multimodal objective promotes the encoding of semantically meaningful nuisances into the image-level representation, $\bs{z}_{\text{cls}+}$, enhancing transfer learning capabilities while trading off linear performance. Table~\ref{table: mmobj} presents the results of linear probing, \emph{k}-NN evaluation, and object detection with different degrees of the multimodal objective involvement during pretraining. We observed that the inclusion of $\mc{L}_{\rm cls+}$ aids in learning enhanced representations at the image level with the regular class token $\bs{z}_{\rm cls}$. This is evident in the more distinct head attention maps shown in Fig.~\ref{fig:attention}. The significant improvement in \emph{k}-NN evaluation suggests that learning a multimodal distribution with distinct modes of nuisances aids in disentangling the information that reflects the image-level distribution of a dataset from the rest of the image. When the multimodal objective is not included, \ie, $\frac{\mc{L}_{\text{cls}+}}{\mc{L}_{\text{cls}}+\mc{L}_p} = 0$, the extra class token functions as a register~\cite{registor}. It brings trivial improvement to the original multi-level objective, as DINO does not exhibit artifact issue.

\vspace{1mm}
\noindent{\bf Blending Factor $\alpha$.} We study the effect of the amount of nuisance information in the multimodal target, controlled by factor $\alpha$. Table~\ref{table: alpha} shows that even a small amount of nuisances brings benefit to the linear performance, suggesting improved semantic information facilitated by the information of image composition.

\begin{table*}[!]
\vspace{-6mm}
\begin{minipage}[l]{0.55\textwidth}
\caption{\footnotesize Effect of semantic constraints and shared projector.}
\label{table: semantics}
\vspace{-4mm}
\resizebox{\textwidth}{!}{
\centering
\begin{tabular}{lcc|c|c}
\toprule[1.pt]
                     
                            $\mc{L}$ Type& 
                            Constraint&
                            SP&
                            Top-1&
                            AP$^{bb}$\\
                            \hline
DINO &  --- & ---  & 74.5 & 40.6\\
DINO &  EsViT\cite{esvit} & \xmark  & 75.3 (+0.8) &  41.8 (+1.2)\\
DINO &  EsViT & \checkmark  & 75.3 (+0.8) &  41.2 (+0.7)\\
DINO &  PixContrast\cite{pixpro} & \xmark & 75.6 (+1.1) & 41.7 (+1.1)  \\
\rowcolor{blue!10} DINO &  UDI &  \xmark & 75.6 (+1.1) &  42.4 (+1.8)\\
\rowcolor{flesh} DINO &  UDI &  \checkmark & 75.8 (+1.3) &  42.6 (+2.0)\\
\bottomrule[1.pt]
\end{tabular}}
\vspace{-1pt}\\
\scriptsize{SP denotes {\it shared projector}. $\dag$ indicates that the method trains equivalently for 300 epochs.}
\end{minipage}
\begin{minipage}[r]{0.4\textwidth}
\caption{\footnotesize Impact of multimodal objective. }
\label{table: mmobj}
\vspace{-4mm}
\resizebox{\textwidth}{!}{
\begin{tabular}{c|ccccc}
\toprule[1.pt]
            $\scaleto{\sfrac{\mc{L}_{\text{cls}+}}{\mc{L}_{\text{cls}}\!+\!\mc{L}_p}}{10pt}$& 
                            \;\;0\;\;&\quad\quad\quad &
                            \;\;0.5\;\;&\quad\quad\quad & 
                            \cellcolor{flesh}\;\;1\;\;\\
                            \hline
Top-1 &  75.8& & 76.2 & & \cellcolor{flesh}76.3 \\
$k$-NN   & 73.0 & &74.1 & & \cellcolor{flesh}74.9 \\
AP$^{bb}$ & 41.8 & & 42.6 & &\cellcolor{flesh}42.5\\
\bottomrule[1.pt]
\end{tabular}}
\vspace{2mm}
\caption{\footnotesize Effect of the blending factor $\bs{\alpha}$.}
\label{table: alpha}
\centering
\vspace{-4mm}
\resizebox{\textwidth}{!}{
\begin{tabular}{c|ccc}
\toprule[1.pt]
                     
                            \quad\quad\;\;$\bs{\alpha}$\;\; \quad\quad & 
                            \quad\;\;\;\;0.1\;\;\;\;\quad\quad &
                            \cellcolor{flesh}\quad\quad\;\;\;0.5\;\;\; \quad\quad&\\
                            \hline
\;\;Top-1\;\; &  75.9 & \cellcolor{flesh} 76.3\\
\bottomrule[1.pt]
\end{tabular}}
\end{minipage}
\end{table*}

\vspace{-10mm}
\section{Conclusion}
\vspace{-2mm}
We address the information compression issues prevalent in current SSL methods and introduce UDI, a novel multi-level SSL approach that learns semantically rich representations by: (i) employing context-aligned semantic constraints via self-attention, (ii) utilizing semantic-type-aligned objectives with a shared projector, and (iii) learning an extra class token \texttt{cls}$+$ to produce multimodal predictions by constructing a target distribution using predictions from different granularities to reflect the image composition. Extensive experiments demonstrate the effectiveness of UDI, which achieves SOTA performance in both classification and dense prediction tasks.

\noindent\textbf{Limitations and broader impacts:} UDI's primary limitation is its current design focus on ViT-based models, with experiments conducted solely on ViT-S/16. Future work could explore extending UDI to non-ViT backbones, and evaluating its performance with larger models and longer training duration. As far as we can foresee, there is no negative societal impacts with UDI.

\section*{Acknowledgement}
Research was sponsored by the Army Research Laboratory and was accomplished under Cooperative Agreement Number W911NF-23-2-0224. The views and conclusions contained in this document are those of the authors and should not be interpreted as representing the official policies, either expressed or implied, of the Army Research Laboratory or the U.S. Government. The U.S. Government is authorized to reproduce and distribute reprints for Government purposes notwithstanding any copyright notation herein.
%
%
\bibliographystyle{splncs04}
\bibliography{egbib}

\newpage
\title{Unsqueeze \texttt{[CLS]} Bottleneck to Learn Rich Representations---Supplementary Material} 

\titlerunning{Unsqueeze \texttt{[CLS]} Bottleneck to Learn Rich Representations}

\author{Qing Su, and Shihao Ji}

\authorrunning{Q. Su, S. Ji}

\institute{Department of Computer Science\\Georgia State University, USA}

\maketitle

\vspace{-5pt}
\section{Related Work}\vspace{-5pt}
A comprehensive review of studies that employ the joint embedding strategy is provided in this section. We categorize these methods into two types based on the nature of their underlying clustering processes: {\bf implicit clustering} and {\bf explicit clustering}.\vspace{2mm}

\noindent\textbf{SSL via implicit clustering}.
Methods based on implicit clustering facilitate clustering without specifying the number of centroids. They achieve this by drawing the embeddings of a sample and its augmented views closer to maximize representation similarity. Building upon this, methods emphasizing dissimilarity through the separation of other samples and their augmentations are categorized under \textbf{contrastive learning} (CL). Notable works in CL, such as those by Misra \& Maaten~\cite{misra2020self}, He et al.~\cite{moco}, Chen et al.~\cite{simclr, mocov2},  Hjelm et al.~\cite{hjelm2018learning},  and Chen et al.~\cite{simclrv2}, largely utilize the InfoNCE loss~\cite{infonce}, achieving enhanced performance with substantial data volumes. However, the indefinite number of clusters introduces a vulnerability to over-fitting within contrastive learning strategies--viewing each sample as its own class. Tschannen et al.~\cite{tschannen2019mutual} demonstrates that InfoNCE-based objectives could be maximized trivially using invertible encoders. Moreover, a fundamental issue with contrastive objectives is their propensity to retain non-essential information as an "identifier", adversely affecting the generalizability of representation~\cite{shwartz2023compress}. \textbf{Non-contrastive} approaches, instead, focus solely on optimizing similarity through the use of cosine similarity loss, as seen in works by Chen \& He~\cite{simsiam}, Grill et al.~\cite{vicreg}, and Bardes et al.~\cite{vicreg}. These methods, devoid of negative samples, employ various techniques to prevent output or embedding collapse. Such strategies include batch-wise or feature-wise normalization, the use of a "momentum encoder" for gradual updates of one branch's parameters with the moving average of another's, the application of a stop-gradient operation in one of the branches~\cite{byol, richemond2020byol}, or covariance regularization to maximize the volume of representation space~\cite{barlowtwins, vicreg}. By mitigating the issues faced in CL, non-contrastive methods have further enhanced performance.

\vspace{2mm}
\noindent\textbf{SSL via explicit clustering}. Another paradigm approaches SSL from the perspective of explicit clustering. DeepCluster~\cite{caron2018deep} generates pseudo-labels for new representations by utilizing the k-means assignments from previous iterations. This process involves an extensive asynchronous clustering phase, which limits the method's scalability. To mitigate this issue, SwAV~\cite{swav} introduces an online learning approach for clustering, while ensuring a balanced partition of assignments through Sinkhorn-Knopp regularization~\cite{cuturi2013sinkhorn}. MSN~\cite{msn} further softens the hard assignment constraint in DeepCluster and SwAV by employing a soft constraint through the entropy of averaged predictions, leading to enhanced performance and efficiency. Incorporating standalone prototypes into the last layer of a prediction head, DINO~\cite{dino} advances knowledge distillation within SSL by  using the prediction vector as a soft label. DINO effectively prevents output collapse with strategies like sharpening and logits centering. By increasing the number of centroids and optimizing the encoder and prediction head end-to-end, DINO achieves notably improved performance, yielding semantically meaningful representations with the vision transformer. Yet, as shown in Figs.~\ref{fig:selfattn1}-\ref{fig:selfattn3}, due to the skewed sharpening treatment in DINO, the method is susceptible to over-compression, leading to information loss.

\vspace{5pt}
\noindent\textbf{Multi-level SSL}. Building on the foundational image-level SSL methods, multi-level SSL frameworks aim to capture a more comprehensive spectrum of information by integrating learning objectives at finer granularities, such as patches and blocks. This approach is designed to enhance the model's understanding of images by focusing on both the global context and the intricate details contained within smaller image segments. For instance, DenseCL~\cite{densecl} innovates by applying contrastive learning to pairs of patches according to their similarity. While this patch-matching technique generates unique representation for each patch, it often results in a low correlation between patches, failing to encapsulate the holistic semantics of natural images fully. To address this issue, PixPro~\cite{pixpro} introduces a mechanism that ensures agreement between positive pixel pairs identified by thresholded proximity. Similarly, ReSim~\cite{resim} seeks to maximize agreement between representations pooled from sliding windows across overlapping regions of two augmented views. Further expanding on this concept, DetCo~\cite{detco} incorporates contrastive losses at both the image-patch and image-level, along with patch-level losses, creating a robust framework that fosters comprehensive representation learning across multiple scales. However, as highlighted in Table~\ref{tab:tradeoff}, directly summing objectives across different levels of granularity can undermine image-level performance, \eg, DenseCL and ReSim. With ViT and its variants as encoder, a series of methods are proposed steming from DINO. EsViT~\cite{esvit}, designed for Swin Transformer~\cite{swin}, follows the region-matching strategy and applies the DINO loss to the probabilities of positive pairs determined by highest similarity. To retain local semantic consistency, SelfPatch~\cite{selfpatch} considers the Top-K similar neighboring patches as its positive patches instead of finding the best-matching patch. Furthermore, iBOT~\cite{ibot} leverages Masked Image Modeling (MIM) to train a student encoder to recover the masked patches. The prediction of  recovered patches and that of original patches from teacher are aligned via a DINO loss. Notably, iBOT enforces alignment of semantic types from image and patch-level through a shared prediction head, disregarding the inherent semantic type constraint imposed through the MIM mechanism. We visualize the attention map of iBOT in Figs.~\ref{fig:selfattn1}-\ref{fig:selfattn3} to illustrate the conflict of semantic information captured by iBOT.

\newpage
\section{Pseudocode}
We provide the PyTorch-like pseudocode of the UDI framework with two-view setting, which can be readily extended to multi-crop setting. For detailed information of data augmentation and stratified random sampling, please refer to Section~\ref{sec:implementation details}.

\vspace{-5mm}
\begin{center}
\begin{minipage}[c]{.9\textwidth}
\begin{algorithm}[H]
\footnotesize
\label{algorithm}
\captionsetup{font=small}
\caption{\textsc{Pytorch Pseudo-code of the UDI Framework}}
   \label{alg:udi}
\begin{algorithmic}[H]
    \scriptsize
   \STATE{\texttt{\color{PineGreen} \# ft, fs: teacher and student vision transformer}}
   \vspace{-2pt}\STATE{\texttt{\color{PineGreen} \# sa\_t, sa\_s: self-attention in SRS module of teacher and student}}
   \vspace{-2pt}\STATE{\texttt{\color{PineGreen} \# tp\_t, tp\_s: student and teacher temperatures}}
   \vspace{-2pt}\STATE{\texttt{\color{PineGreen} \# ht, hs: projectors for teacher and student branches}}
   \vspace{-2pt}\STATE{\texttt{\color{PineGreen} \# $\alpha$: blending factor}}
   \vspace{-2pt}\STATE{\texttt{\color{PineGreen} \# $\beta_1$, $\beta_2$: network and prediction center momentum rates}}
   \vspace{-2pt}\STATE{\texttt{\color{PineGreen} \# UdiTransforms: data transform with sampling masks generated}}
   \STATE
   \STATE \texttt{{\bfseries def} H(t, s):\texttt{\color{PineGreen}\# cross-entropy loss}}
   \STATE\hspace{10pt} \texttt{{\bfseries return} - (t * log(s)).sum(dim=-1).mean()}
   \STATE
   \STATE \texttt{{\bfseries def} CS(x, C=None):\texttt{\color{PineGreen}\# centering and sharpening}}
   \STATE\hspace{10pt} \texttt{if C is not None:}
   \STATE\hspace{20pt} \texttt{{\bfseries return} softmax((x - C) / tp\_t, dim=-1)}
   \STATE\hspace{10pt} \texttt{else:}
   \STATE\hspace{20pt} \texttt{{\bfseries return} softmax(x / tp\_s, dim=-1)}
   \STATE
   \STATE \texttt{ft.params = fs.params}
   \STATE \texttt{{\bfseries for} x {\bfseries in} Loader:}{\texttt{\color{PineGreen}\# load a minibatch of x with B samples}}
   \STATE\hspace{10pt}{\texttt{\color{PineGreen}\# random augmentations and  random sampling masks}}
   \STATE \hspace{10pt} \texttt{x1, x2, mask1, mask2 = UdiTransforms(x)}
   \STATE \hspace{10pt} \texttt{S1, T2 = fs(x1), ft(x2)}{\texttt{\color{PineGreen}\;\;\# [B, N, D]}}
   \STATE \hspace{10pt} \texttt{s1, e\_s1, S1 = S1[0], S1[1], S1[2:]}
   \STATE \hspace{10pt} \texttt{t2, T2 = T2[0], T2[2:]}
   \STATE \hspace{10pt} \texttt{patch\_s1 = sa\_s(S1[mask1], S1, S1)}
   \STATE \hspace{10pt} \texttt{patch\_t2 = sa\_t(T2[mask2], T2, T2)}
   \STATE
   \STATE \hspace{10pt} {\texttt{\color{PineGreen}\# image-level output [B, K]}}
   \STATE \hspace{10pt} \texttt{v\_s1, v\_se1, v\_t2 = hs(s1), hs(e\_s1), ht(t2)}
   \STATE \hspace{10pt} {\texttt{\color{PineGreen}\# patch-level output [B, L, K]}}
   \STATE \hspace{10pt} \texttt{v\_ps1, v\_pt2 = hs(patch\_s1), ht(patch\_t2)}
   \STATE \hspace{10pt} \texttt{\color{PineGreen}\# centering and sharpening}
   \STATE \hspace{10pt} \texttt{pred\_s1, pred\_se1, pred\_ps1 = CS(v\_s1), CS(v\_se1), CS(v\_ps1)}
   \STATE \hspace{10pt} \texttt{pred\_t2, pred\_pt2 = CS(v\_t2.detach(), Ci), CS(v\_pt2.detach(), Cp)}
   \STATE \hspace{10pt} \texttt{\color{PineGreen}\# target multi-modal distribution}
   \STATE \hspace{10pt} \texttt{pred\_mm = $\alpha$ * pred\_pt2.mean(dim=1) + (1 - $\alpha$) * pred\_t2}
   
   \STATE
   \STATE \hspace{10pt} \texttt{\color{PineGreen}\# image-level, patch-level, multi-modal(mm) loss:}
   \STATE \hspace{10pt} \texttt{loss\_image = H(pred\_t2, pred\_s1)}
   \STATE \hspace{10pt} \texttt{loss\_patch = H(pred\_pt2, pred\_ps1)}
   \STATE \hspace{10pt} \texttt{loss\_mm = H(pred\_mm, pred\_se1)}
   \STATE \hspace{10pt} \texttt{loss = loss\_image + loss\_patch + loss\_mm}
   \STATE \hspace{10pt} \texttt{loss.backward()}{\texttt{\color{PineGreen}\;\;\# back-propagation}}
   
   \STATE
   \STATE \hspace{10pt} \texttt{\color{PineGreen}\# student and teacher updates}
   \STATE\hspace{10pt} \texttt{updates(fs)}{\texttt{\color{PineGreen}\;\; \# SGD}}
   \STATE\hspace{10pt} \texttt{ft.params = $\beta_1$ * ft.params + (1-$\beta_1$) * fs.params}
   \STATE\hspace{10pt} \texttt{Ci= $\beta_2$ * Ci + (1-$\beta_2$) * v\_t2.mean(dim=0)}
   \STATE\hspace{10pt} \texttt{Cp= $\beta_2$ * Cp + (1-$\beta_2$) * v\_pt2.mean(dim=(0,1))}
\end{algorithmic}
\end{algorithm}
\end{minipage}
\end{center}

\newpage
\section{Implementation Details}\label{sec:implementation details}

Owning to space constraints in the main paper, we provide additional experimental details in this section. Next, we expand implementation details on pretraining, followed by the experimental setup for various downstream tasks.\vspace{2mm}

\noindent\textbf{Augmentation}. We largely follow the data augmentation techniques used in BYOL~\cite{byol} and DINO ~\cite{dino}, which are widely adopted by many SSL works. The augmentation includes random crop, color jitter, Gaussian noise, gray scaling, and horizontal flipping. We use the same hyperparameters for those augmentation operations as utilized in DINO~\cite{dino}.

\noindent\textbf{Multi-crop setting}. 
Following conventional multi-crop setting~\cite{swav,dino}, we crop each image into 2 large crops of size 224 and 10 extra small crops of size 96. To ensure the correspondence for patch-level representations. Specifically, we first crop the student's global view with a scale range of $[0.4, 1]$ and subsequently crop the teacher's global view with a minimum overlap ratio ($\geq$0.25) using the same scale.  For local crops, we apply a scale of $[0.05, 0.4]$ relative to the original image with a minimum overlap ratio with the intersection of the two global views from teacher and student models. 

To calculate the total loss of Eq.~11 for multiple crops, we regard one global crop as $\bs{X}_1$ and take the remaining 11 crops as $\bs{X}_2$. Similarly, we treat another global crops as $\bs{X}_1$ and the remaining 11 crops as $\bs{X}_2$. The losses derived from these two configurations are averaged to compute the overall training loss.\vspace{2mm}

\noindent\textbf{Pretraining hyperparameter settings}.
We pretrain ViT-S/16 on the ImageNet-1K dataset without labels, utilizing the AdamW optimizer~\cite{adamw} and a minibatch size of 1024. Consistent with the DINO framework, the learning rate undergoes a linear warm-up during the initial 10 epochs from $10^{-6}$ to its base value of $8\times10^{-4}$, as prescribed by the linear scaling rule~\cite{goyal2017accurate}, e.g., $2\times10^{-4}/256$. Following this warm-up phase, we apply a cosine scheduler~\cite{cosine} for learning rate decay with a final learning rate of $10^{-6}$. Similarly, weight decay is modulated by a cosine scheduler, transitioning from 0.04 to 0.1. In all the experiments, the learning rate for the patch embedding layer is set to be $5\times$ lower than the base rate, aligning with MoCo-v3~\cite{mocov3} to enhance training stability. Like DINO, we adopt a drop path rate of 0.1 and set the gradient clipping threshold to 3.0 for ViT-S/16. The student temperature $\tau_s$ is fixed at 0.1, whereas the teacher temperature $\tau_t$ undergoes a linear warm-up from 0.04 to 0.07. The majority of UDI's configurations closely follow those of DINO in order to simplify the setup, minimize hyperparameter adjustment, and reduce computational costs.\vspace{2mm}

\noindent{\bf Training cost of UDI.} Table~\ref{table: train_req} reports the training time and memory cost of UDI, DINO and iBOT. Although UDI learns an extra class token and an SA module, it samples less local patches (8\%) than iBOT ($\sim$15\%). Thus, the training times and memory costs of UDI is similar to that of iBOT.\vspace{2mm}
\begin{table}[h]
\centering
\begin{minipage}[h]{\linewidth}
\centering
\caption{Time and Memory Requirements on A100 nodes}
\label{table: train_req}\vspace{-5pt}
\resizebox{.6\textwidth}{!}{
\begin{tabular}{c|cccc}
\toprule[1.pt]
                            \;\;\;Method\;\;\;& 
                            \#crops&
                            \;\;\;$T_{100}$\;\;\;&
                            \;\;\;$T_{300}$\;\;\;&
                            \;\;\;Mem.\;\;\;\\
                            \hline
DINO & $2\times 224^2 + 10\times 96^2$ & 22.4h &67.3h & 15.4G\\
iBOT & $2\times 224^2 + 10\times 96^2$ & 22.6h &67.8h & 19.5G\\
UDI & $2\times 224^2 + 10\times 96^2$ & 23.1h &69.3h & 19.8G\\
\bottomrule[1.pt]
\end{tabular}}
\end{minipage}\vspace{-5pt}
\end{table}

\noindent\textbf{Low-shot learning}.
We adopt two evaluation approaches: 1) train a logistic regression classifier on top of frozen features, and 2) end-to-end fine-tune the entire pretrained backbone. For logistic regression classifier, we extract the frozen feature as in \emph{k}-NN evaluation. Following DINO, we train the classifier for 100 epochs using the AdamW optimizer with a minibatch size of 1024 under both 1\% and 10\% data. For regularization parameter, we sweep over $\{0.01, 0.1, 1.0\}$. To fine-tune the whole pretrained model, we adopts iBOT settings for ViT-S/16, keeping the first layer of the projection head. The model is fine-tuned with 1000 epochs using AdamW with a minibatch size of 1024 and weight decay 0.05 under both 1\% and 10\% training data settings with a learning rate of $5\times10^{-6}$.\vspace{2mm}

\noindent\textbf{Transfer learning}. Following the training recipe and protocol adopted by iBOT, we fine-tune an ImageNet-1K pretrained ViT-S/16 on several smaller datasets. We employ the AdamW optimizer across all the experiments, utilizing a minibatch size of 768. The model is trained for 1000 epochs on the CIFAR10, CIFAR100, Cars, and Flwr datasets. A uniform learning rate of $7.5\times10^{-6}$ is applied to all these datasets, with the exception of the Cars dataset, to which a higher learning rate of $1\times10^{-4}$ is used. For iNat18 and iNat19, training durations are set to 360 epochs, with learning rates of $2.5\times10^{-5}$ and $7.5 \times 10^{-5}$, respectively.

\vspace{2mm}
\noindent\textbf{Object detection \& Instance segmentation}. 
 We evaluate the performance of the pretrained ViT-S/16 on MS-COCO object detection and instance segmentation tasks utilizing a two-staged detection framework Mask R-CNN~\cite{maskrcnn} with FPN~\cite{fpn}. Following~\cite{carion2020end}, we adopt multi-scale training and adjust image sizes so that the short side is between 480 and 800 pixels, with the long side capped at 1333 pixels. In line with the training configurations specified in~\cite{selfpatch}, our experiments are conducted using the AdamW optimizer with a batch size of 8. The learning rate experiences a linear warm-up over the initial 1000 iterations to reach $5\times10^{-5}$, followed by step decay by 10 times at steps 8 and 11. For a standardized comparison, all models reported in Table 4 are trained with 1x schedule.\vspace{2mm}

\noindent\textbf{Semantic segmentation}.
In line with Selfpatch~\cite{selfpatch}, we follow the configurations of MMSegmentation~\cite{mmseg2020} to fine-tune Semantic FPN and UPerNet framework for 40K and 160K iterations, respectively. Both frameworks are trained with input resolution of $512\times512$ and the AdamW optimizer~\cite{adamw}. The learning rate undergoes a linear warm-up for the first 1500 iterations to reach the base value $6\!\times\!10^{-5}$ followed by linear learning rate decay. Weight decay is set to 0.01, excluding positional embedding, class tokens, and layer norm.

\begin{table*}[th!]
\begin{minipage}[l]{\textwidth}
\vspace{2.5mm}
\caption{ImageNet-1K linear evaluation and MS-COCO object detection performance of the methods adopting multi-level clustering pretext and their corresponding instance-level fundamental methods (denoted in gray background).}\label{tab:tradeoff}
\vspace{-3mm}
\centering
\resizebox{0.75\textwidth}{!}{\begin{tabular}{lccccc}
    \toprule[1.pt]
    Objective & Arch. & \#Views & Epoch$^{\dag}$ & Linear &AP$^{bb}$\\
    \hline
    \rowcolor{lgray} MoCov2  & RN50 & 2 & 400 & 67.5 & 38.9\\
    \ \ \ DenseCL~\cite{densecl} & RN50 & 2 & 400 & 64.6~\textcolor{red}{($-$2.9)}& 40.3~\textcolor{blue}{(+1.4)} \\
    \ \ \ ReSim~\cite{resim}  & RN50 & 2 & 400 & 66.1~\textcolor{red}{($-$1.4)} &40.3~\textcolor{blue}{(+1.4)}\\
    \ \ \ DetCo~\cite{detco}  & RN50 & 2 & 400 & 68.6~\textcolor{blue}{(+1.1)} & 40.1~\textcolor{blue}{(+1.2)}\\
    \rowcolor{lgray} SimCLR~\cite{simclr}  & RN50 & 2 & 200 & 65.4 &40.5\\
    \ \ \ PixPro~\cite{simclr}  & RN50 & 2 & 200 & 66.3~\textcolor{blue}{(+0.9)} &40.9~\textcolor{blue}{(+0.5)} \\
    \rowcolor{lgray} DINO~\cite{dino}   & ViT-S/16 & 10 & 1050 & 76.0 &41.5\\
    \ \ \ Selfpatch~\cite{selfpatch}   & ViT-S/16 & 10 & 1050 & 75.6~\textcolor{red}{(-0.4)} & 42.1~\textcolor{blue}{(+0.6)}\\
    \rowcolor{lgray} DINO~\cite{dino}   & Swin-T/14 & 10 & 1050 & 77.1 &46.0\\
    \ \ \ EsViT~\cite{esvit}   & Swin-T/14 & 10 & 1050 & 77.6~\textcolor{blue}{(+0.5)}& 46.2 ~\textcolor{blue}{(+0.2)}\\
    \rowcolor{lgray} DINO~\cite{dino}   & ViT-S/16 & 12 & 1200 & 76.1 &41.6\\
    \ \ \ {iBOT}~\cite{ibot}   & ViT-S/16 & 12 & 1200 & 77.4~\textcolor{blue}{(+1.3)} &41.7~\textcolor{blue}{($+$0.1)}\\
    \rowcolor{lgray} DINO~\cite{dino}   & ViT-S/16 & 12 & 1200 & 76.1 &41.6\\
    \ \ \ {\bf UDI}   & ViT-S/16 & 12 & 1200 & 77.6~\textcolor{blue}{(+1.5)} & 43.2~\textcolor{blue}{(+1.6)}\\
\bottomrule[1.pt]
\end{tabular}}
\vspace{5mm}
\end{minipage}

\begin{minipage}[c]{\textwidth}
    \caption{Object detection and instance segmentation on MS-COCO with Mask R-CNN under 1x schedule. Epoch refers to the number of pretraining epochs on ImageNet-1K.}
    \vspace{-8pt}
    \footnotesize\centering
    \resizebox{\textwidth}{!}{\begin{tabular}{l|l|ccc|cccccc}
    \toprule[1.pt]
        ~ & {Method} & {Backbone} & \#Params. & {\#Epo.} & AP$^{bb}$ & AP$^{bb}_{50}$ & AP$^{bb}_{75}$ & AP$^{mk}$ & AP$^{mk}_{50}$ & AP$^{mk}_{75}$\\
        \hline
        \multirow{6}{*}{\rotatebox[origin=c]{90}{\makebox[0.25\columnwidth]{\scalebox{0.9}{\hspace{8mm}Image-level SSLs}}}} & MoCo-v2~\cite{mocov2}  &RN50  &23M  &200  &38.9  &59.2 &42.4 &35.5 &56.2 &37.8\\
        & SwAV~\cite{swav} & RN50 &23M & 200 & 38.5 & 60.4 & 41.4 & 35.4 & 57.0 & 37.7 \\
        & MOCO-v3~\cite{mocov3}  &ViT-S/16  &21M  &300  &39.8   &62.6  &43.1   &37.1  &59.6  &39.2\\
        & MoBY~\cite{moby}  &ViT-S/16 &21M  &300  &41.1  &63.7  &44.8  &37.6  &60.3  &39.8 \\
        & DINO~\cite{dino}  &ViT-S/16  &21M  &800  &40.8   &63.6  &44.2  &37.4  &60.1  &39.5 \\
        & DINO300-lc6~\cite{dino}  &ViT-S/16  &21M  &300  &41.5   &64.1  &45.1  &37.8  &60.5  &40.0 \\
        & DINO300-lc10~\cite{dino}  &ViT-S/16  &21M  &300  &41.6   &64.4  &45.2  &38.0  &60.9  &40.3 \\
        & MSN~\cite{msn} & ViT-S/16 & 21M & 800 &41.7 &64.6 & 45.1  &38.1 & 61.2 & 40.5 \\	
        \hline
        \multirow{6}{*}{\rotatebox[origin=c]{90}{\makebox[0.25\columnwidth]{\scalebox{0.9}{\hspace{8mm}Multi-level SSLs}}}} &DenseCL~\cite{densecl}  &RN50   &23M &200 &40.3  &59.9 &44.3 &36.4 &57.0 &39.2\\
        & ReSim~\cite{resim} & RN50 &23M & 200 & 40.3 &60.6 & 44.2  & 36.4 & 57.5 & 38.9\\
        & DetCo  &RN50  &23M &200   &40.1  &61.0 &43.9 &36.4 &58.0  &38.9\\
        & iBOT~\cite{ibot} & ViT-S/16 &21M & 800 &41.7 & 64.1 &45.5 &38.1 &60.8 &40.6\\
        & DINO+SelfPatch~\cite{selfpatch}  &ViT-S/16    &21M  &200  &42.1  &64.9  &46.1  &38.5  &61.3  &40.8\\
        & \cellcolor{flesh}UDI & \cellcolor{flesh}ViT-S/16 & \cellcolor{flesh}21M & \cellcolor{flesh}300 & \cellcolor{flesh}{\bf 43.2} &\cellcolor{flesh}{\bf 66.0} &\cellcolor{flesh}{\bf 47.7}  &\cellcolor{flesh}{\bf 39.4} &\cellcolor{flesh}{\bf 62.4} &\cellcolor{flesh}{\bf 41.7}  \\			
        \bottomrule[1.pt]
    \end{tabular}}
    \label{tab:coco_det_seg_full}
\end{minipage}
\vspace{-2mm}
\end{table*}

\section{More Results and Visualizations}
In this section, we further discuss the impact of multi-level objectives on image classification and dense prediction tasks, followed by the complete results of MS-COCO object detection and instance segmentation. We conclude by showcasing attention visualizations from various notable methods, offering insights into the effectiveness of the UDI framework.
\vspace{-2mm}
\subsection{Effectivenss of multi-level objectives}
We compare the UDI objective against the existing multi-level objectives, including DenseCL~\cite{densecl}, DetCo ~\cite{detco}, PixPro~\cite{pixpro}, EsViT~\cite{esvit}, and iBOT~\cite{ibot}. Most of these objectives improve the performance in object detection as well as ImageNet classification. Among them, UDI achieves the best trade-off: improving the performance of both tasks, with a sophisticated multi-level, global-local interaction algorithm. 
\vspace{-2mm}
\subsection{Complete result of object detection and instance segmentation}
Please refer to Table~\ref{tab:coco_det_seg_full}
\vspace{-2mm}
\subsection{More attention visualization results}
Similar to Fig.~4. in the main paper, here we visualize more attention maps from the final layer of a UDI pretrained ViT-S/16, alongside comparisons with visualizations from two key methods based on self-distillation, \ie, DINO and iBOT. Beyond traditional head attention, we visualize the class token self-attention (\texttt{cls} probing) as an averaged information response, which helps highlight the quantity and type of information encapsulated within the class token. To demonstrate the distinct characteristics of each method more effectively, we present the full attention map rather than restricting our view to the top $\kappa$ percent of the mass (\eg, $\kappa$=60\%). 

In Figs~\ref{fig:selfattn1}-\ref{fig:selfattn3}, we use UDI and UDI$_+$ to represent visualizations based on the standard class token $\bs{z}_{\text{cls}}$ and the extra class token $\bs{z}_{\text{cls}+}$, respectively. Meanwhile, DINO300 and DINO800 indicate the models pretrained with DINO for 300 and 800 epochs, respectively. The visualizations of \texttt{cls} probing and head attention reveal that the UDI framework produces attention maps closely aligned with semantic content of images. While UDI tends to emphasize foreground objects more, UDI$_+$ provides detailed attention to both foreground and background elements. This differentiation suggests that $\bs{z}_{\text{cls}}$ captures more concentrated information that mirrors the image-level distribution across the dataset, whereas $\bs{z}_{\text{cls}+}$ maintains a less condensed form of data, preserving significant nuances that capture both the local context and main subjects. 

In addition, DINO800 generates "cleaner" attention maps than DINO300 by primarily omitting most of the background details. This effect aligns with the "over-compression" issue highlighted in the main paper, where the model is driven to produce progressively more focused predictions throughout training. While the more condensed representation from DINO800 results in enhanced linear performance, indicated by its 77.0\% versus DINO300's 76.1\% as noted in~\cite{dino}, this efficiency trades off its effectiveness in dense prediction tasks, exemplified by its 40.3 AP$^{bb}$ compared to DINO300's 41.1 AP$^{bb}$ in object detection.

On the other hand, iBOT generates more focused and disentangled attention map between foreground and background, largely attributed to its Masked Image Modeling (MIM) objective and the use of a shared prediction head between patch and class tokens. As a result, the attention per head is more attracted to semantics at patch-level, which is evident in its head attention map.

Furthermore, the visualization of head attention reveals that distinct heads within UDI and UDI$_+$ focus on varied semantic elements of an image, indicating a diversification in attention. In contrast, the attention maps produced by DINO display more uniformity, with a higher degree of correlation observed across its different heads. 

\vspace{-3mm}
\section{More Ablations}\vspace{-2mm}
\textbf{Stratified Random Sampling}.
We investigate the impact of stratified random sampling of patches on the construction of multi-modal target predictions and the computation of patch-level loss. This stratified random sampling operates as a form of random pooling within a specified window. Through experiments with various window sizes, we analyze the linear performance of the ViT-S/16 model, trained for 100 epochs. As shown in Table~\ref{tab:window_size}, utilizing a window size of $3\times3$—approximately 10\% of the total patches—results in the best performance. Conversely, employing the entire set of patches (equivalent to a window size of 1) leads to only marginal improvements. We observe that using all patches tends to produce an over-smoothed compound posterior as outlined in Eq.~5, causing the student model to essentially learn from a noisy regular teacher prediction with the extra class token, and thus, only achieving limited enhancement. Additionally, when comparing with a regular random pooling method, \eg, uniform random pooling within an image, stratified random sampling excels by offering better training stability and superior performance. \vspace{-5mm}

\begin{table*}[h!]
    \begin{minipage}[b]{\textwidth}
\caption{\footnotesize Effect of window size for stratified random sampling.}
\label{tab:window_size}
\vspace{-2mm}
\centering
\resizebox{0.6\textwidth}{!}{
\begin{tabular}{c|cccc}
\toprule[1.pt]
                     
                            \;\;Window size\;\; & 
                            \;\;\;$1\times1$\;\;\; 
                            & \;\;\;$2\times2$\;\;\; &\cellcolor{flesh}\;\;\;$3\times3$\;\;\; &\;\;\;$4\times4$\;\;\ \\
                            \hline
\;\;Top-1\;\; & 75.5 & 76.2& \cellcolor{flesh} 76.3 & 75.9\\
\bottomrule[1.pt]
\end{tabular}}
\end{minipage}
\vspace{-3mm}
\end{table*}

\vspace{-3mm}
\noindent{\bf Complete MIB Objective.} To further highlight the advantages of UDI, we experiment with an objective constructed from the complete MIB by augmenting the DINO loss with the InfoNCE loss~\cite{simclr}, as shown below:
\vspace{-2mm}
\begin{equation}
    \scaleto{\mc{L}_{MIB} = \mathbb{E}_{\bs{z}, \bs{z}'\in \mathcal{Z}}\left[-\log\left(\frac{\exp(\bs{z}^\top \bs{z}')}{\sum_{\bs{z}_j\in\mathcal{Z}}\exp(\bs{z}^\top \bs{z}_j)}\right) + \beta \text{H}(p(y|\bs{z}'), q(y|\bs{z}))\right]}{25pt},
    \vspace{-2mm}
\end{equation}
where the symmetric KL divergence in Eq. 1 of the main paper is first simplified to the forward KL (due to the distillation approach) and further reduced to the cross-entropy loss as in DINO. Table~\ref{table: MIB} reports the evaluation results with different loss ratios $1/\beta$. A ratio of $1/\beta = 0$ corresponds to the DINO loss, while $1/\beta = \infty$ corresponds to InfoNCE. The results show that the inclusion of InfoNCE leads to limited improvement, saturating around $\beta=1$. This suggests that InfoMAX is not as effective as UDI in preserving semantically significant nuisances.
\vspace{-5mm}

\begin{table*}[h]
\begin{minipage}[h]{\textwidth}
\caption{\footnotesize Effectiveness of the complete MIB objective.}
\label{table: MIB}
\vspace{-3mm}
\centering
\resizebox{0.65\textwidth}{!}{
\begin{tabular}{c|cccc}
\toprule[1.pt]     
                            \;\;\;$\frac{SimCLR}{DINO}=\sfrac{1}{\beta}$\;\;\;& 
                            \;\;\;\;\;\;\;0\;\;\;\;\;\;\; &
                            \;\;\;\;\;\;\;0.5\;\;\;\;\;\;\;&
                            1&
                            \;\;\;\;\;\;\;$\infty$\;\;\;\;\;\;\;
                            \\
                            \hline
Top-1 & 74.5 & 75.1 & 75.3 & 71.9\\
AP$^{bb}$ & 40.6 & 41.3 & 41.4& 39.8\\
\bottomrule[1.pt]
\end{tabular}}
\end{minipage}
\end{table*}

\newpage
\renewcommand{\thefigure}{1.\arabic{figure}}
\begin{figure}[h!]
  \centering
  \includegraphics[width=\textwidth]{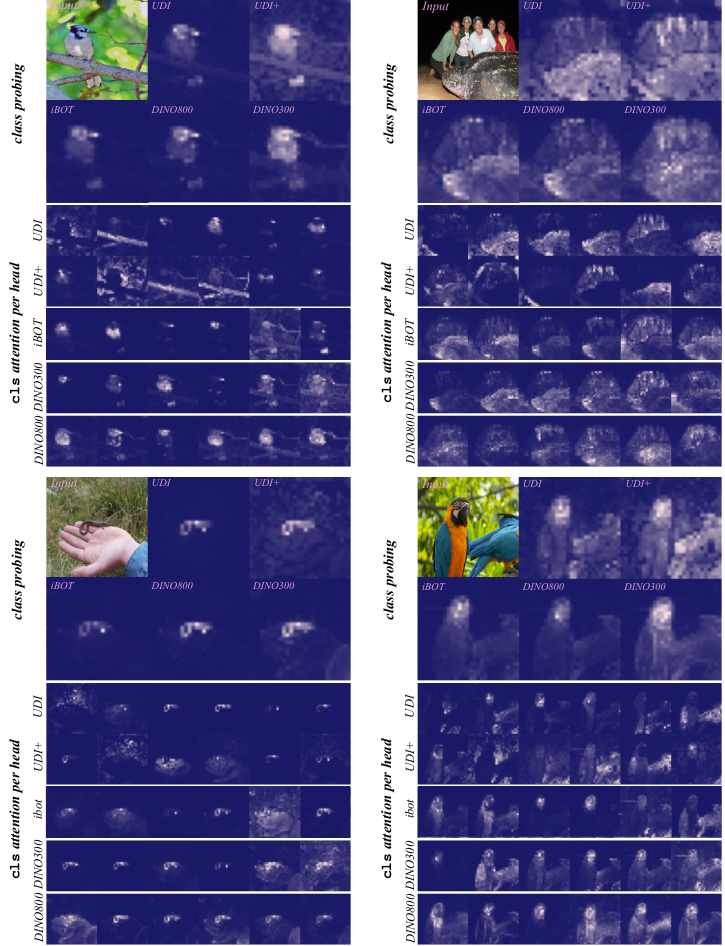}\vspace{-2mm}
  \caption{\footnotesize{\bf Self-attention visualization of ViT-S/16 pretrained with UDI.} UDI and UDI$_+$ denote the visualization using standard and extra class tokens ($\bs{z}_{\text{cls}}$ and $\bs{z}_{\text{cls}+}$, respectively). DINO300 and DINO800 refer to models pretrained with DINO for 300 and 800 epochs, respectively. The UDI framework generates attention maps closely aligned with semantic content of images, where $\bs{z}_{\rm cls}$ focuses more on foreground objects and $\bs{z}_{\rm cls+}$ details both foreground and background elements.}
  \label{fig:selfattn1}
\end{figure}
\newpage
\begin{figure}[h!]
  \centering
  \includegraphics[width=\textwidth]{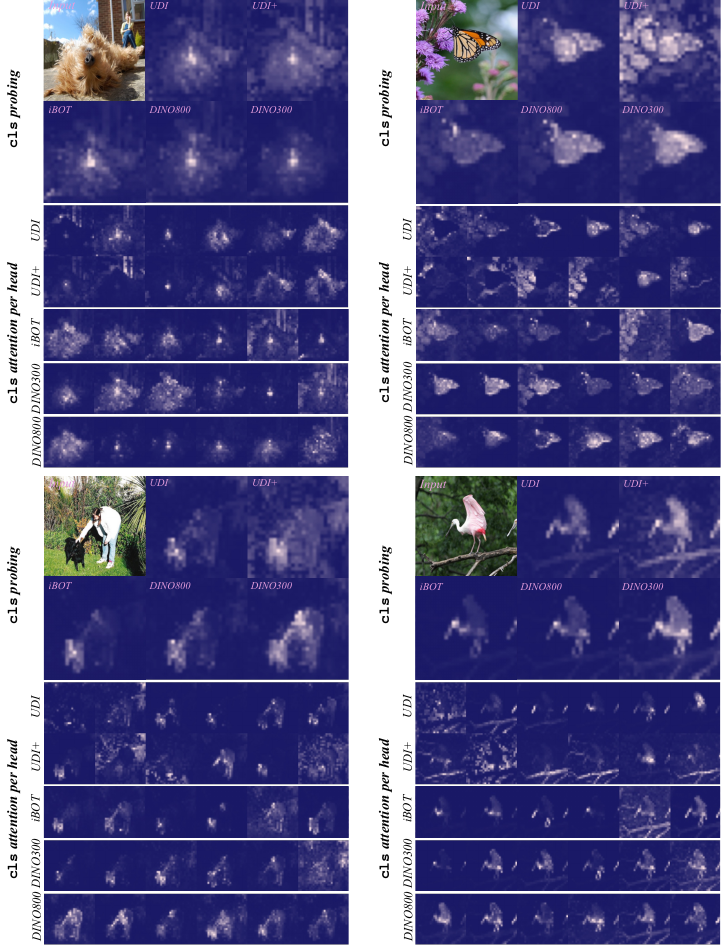}\vspace{-2mm}
  \caption{
  \footnotesize{\bf Self-attention visualization of ViT-S/16 pretrained with UDI.} 
  }
  \label{fig:selfattn2}
\end{figure}
\newpage
\begin{figure}[h!]
  \centering
  \includegraphics[width=\textwidth]{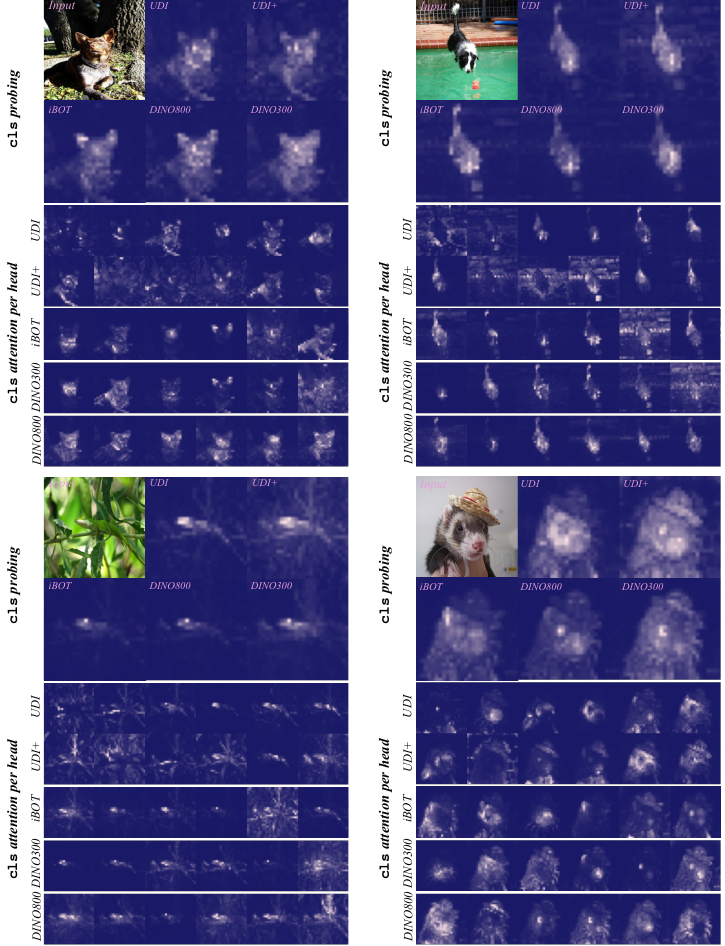}\vspace{-2mm}
  \caption{
  \footnotesize{\bf Self-attention visualization of ViT-S/16 pretrained with UDI.} 
  }
  \label{fig:selfattn3}
\end{figure}

\end{document}